\documentclass[10pt,twocolumn,letterpaper]{article}

\usepackage{iccv}
\usepackage{times}
\usepackage{epsfig}
\usepackage{graphicx}
\usepackage{amsmath}
\usepackage{amssymb}
\usepackage{booktabs}

\usepackage{graphicx}
\usepackage{caption}
\usepackage{subcaption}
\usepackage{amsmath}
\usepackage{amsfonts}
\usepackage{booktabs}
\usepackage{siunitx}
\usepackage{wrapfig}
\usepackage{bbm}
\usepackage{multirow}
\usepackage{relsize}
\usepackage{color}
\usepackage{colortbl}
\usepackage{algorithm}
\usepackage{listings}

\usepackage{enumitem}
\usepackage{url}

\usepackage[accsupp]{axessibility}  %

\definecolor{almond}{rgb}{0.94, 0.87, 0.8}

\newcommand{\myparagraph}[1]{\vspace{2pt}\noindent{\bf #1}}

\usepackage[format=plain,labelformat=simple,labelsep=period,font=small,compatibility=false]{caption}
\usepackage[font=footnotesize,skip=3pt,subrefformat=parens]{subcaption}

\usepackage[pagebackref=true,breaklinks=true,letterpaper=true,colorlinks,bookmarks=false]{hyperref}

\iccvfinalcopy %

\def\iccvPaperID{7411} %

\ificcvfinal\fi

\begin{document}

\title{Learning by Sorting: Self-supervised Learning\\ with Group Ordering Constraints}

\author{%
    Nina Shvetsova$^{1,2,3}$ \quad
    Felix Petersen$^4$  \quad
    Anna Kukleva$^2$  \quad
    Bernt Schiele$^{2}$  \quad
    Hilde Kuehne$^{1,3,5}$ \\
    \kern-1.1em\small{
    $^1$Goethe University Frankfurt,   
    $^2$Max-Planck-Institute for Informatics,
    $^3$University of Bonn,
    $^4$Stanford University, 
    $^5$MIT-IBM Watson AI Lab
    } \\
    \small{
    \texttt{\{shvetsov@uni-frankfurt.de, mail@felix-petersen.de\}}}  
}

\maketitle
\ificcvfinal\thispagestyle{empty}\fi

\begin{abstract}

Contrastive learning has become an important tool in learning representations from unlabeled data mainly relying on the idea of minimizing distance between positive data pairs, e.g., views from the same images, and maximizing distance between negative data pairs, e.g., views from different images. 
This paper proposes a new variation of the contrastive learning objective, Group Ordering Constraints (GroCo), that leverages the idea of sorting the distances of positive and negative pairs and computing the respective loss based on how many positive pairs have a larger distance than the negative pairs, and thus are not ordered correctly. 
To this end, the GroCo loss is based on differentiable sorting networks, which enable training with sorting supervision by matching a differentiable permutation matrix, which is produced by sorting a given set of scores, to a respective ground truth permutation matrix.
Applying this idea to groupwise pre-ordered inputs of multiple positive and negative pairs allows introducing the GroCo loss with implicit emphasis on strong positives and negatives, leading to better optimization of the local neighborhood.
We evaluate the proposed formulation on various self-supervised learning benchmarks and show that it not only leads to improved results compared to vanilla contrastive learning but also shows competitive performance to comparable methods in linear probing and outperforms current methods in $k$-NN performance.
\footnote{\url{{https://github.com/ninatu/learning_by_sorting}}\\ To be published at ICCV 2023. Cite as:
Nina Shvetsova, Felix Petersen, Anna Kukleva, Bernt Schiele, Hilde Kuehne. 
``Learning by Sorting: Self-supervised Learning with Group Ordering Constraints''.
In: \textit{Proceedings of the IEEE/CVF International Conference on Computer Vision}, 2023.
\vspace{-3em}
}

\end{abstract}
\vspace{-.5em}

\vspace{-.75em}
\section{Introduction}

Self-supervised learning has become a topic of growing interest over the last years as it allows models to learn representations from large-scale data without the need for human annotation.
Many approaches rely on the idea of contrastive learning and were able not only to narrow the gap to the supervised learning performance in vision~\cite{dwibedi2021little,chen2020simple,tian2020makes,he2020moco,bachman2019learning,ye2019unsupervised}, but also to train state-of-the-art vision-language~\cite{Radford2021CLIP,schuhmann2022laionb} and multimodal models~\cite{miech20endtoend}.
All of these methods rely on the concept of the pairwise contrastive loss, which is based on the idea that a so-called positive pair, e.g., an image serving as an anchor and an augmentation of the same image, should be closer to each other in an embedding space than a so-called negative pair, e.g., a pair made up of an anchor image and a different image, should be far away from each other. 
\begin{figure}[]
    \centering 
        \includegraphics[width=0.9\linewidth]{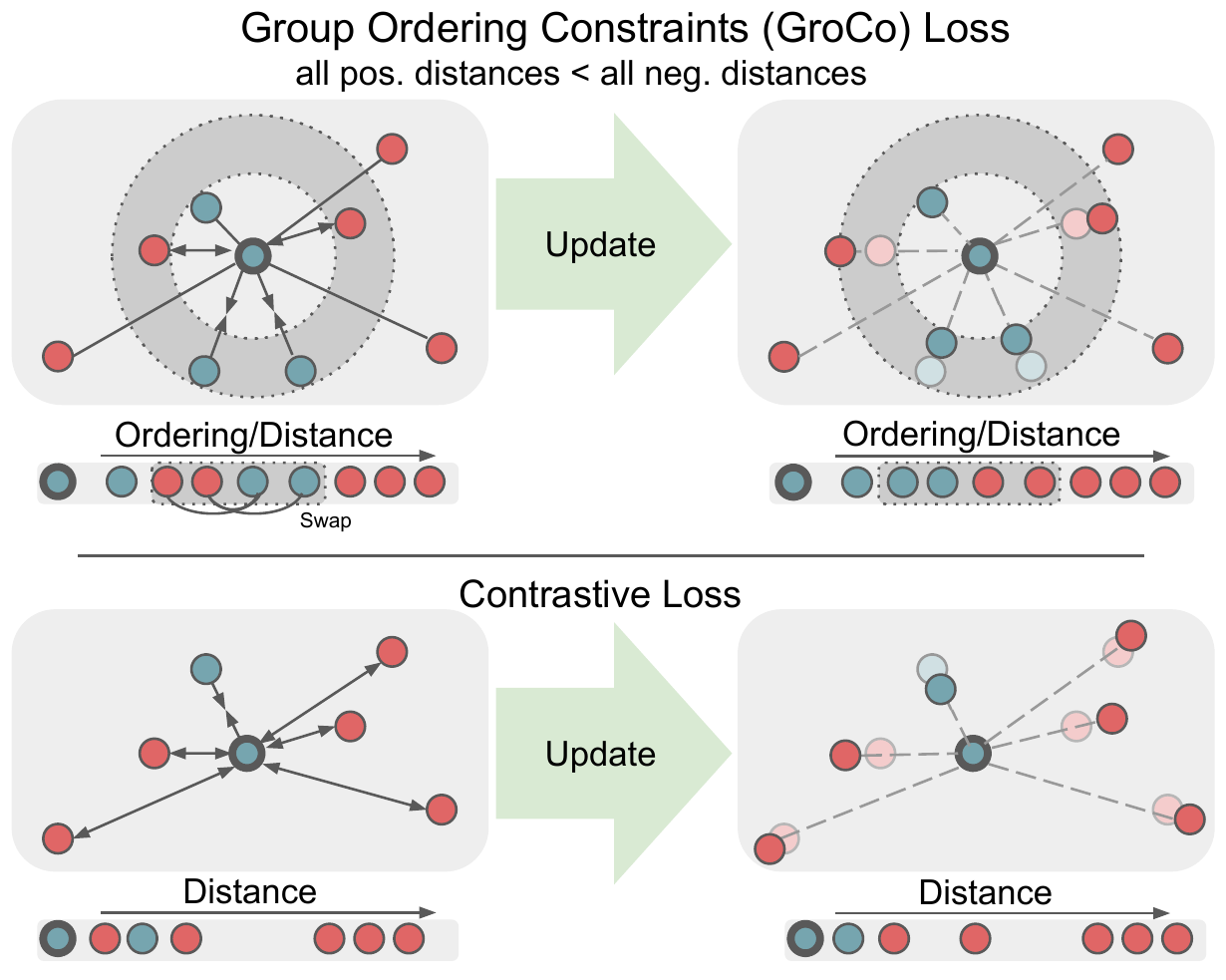}
     \vspace{-0.1cm}
    \caption{
    The idea of the proposed group ordering constraints loss compared to pairwise contrastive losses: GroCo arranges positive and negative data points so that the largest distance to positives must be smaller than the smallest distance to negative points. 
    To this end, the loss implicitly minimizes the amount of necessary swap operation to achieve the ordering constraint. 
    Thus, it focuses on overlapping positives and negatives compared to standard contrastive losses that minimize resp.\ maximize all pairwise distances. 
    \label{fig:idea}}
    \vspace{-2em}
\end{figure}

\begin{figure*}[]
    \centering 
        \includegraphics[width=\textwidth]{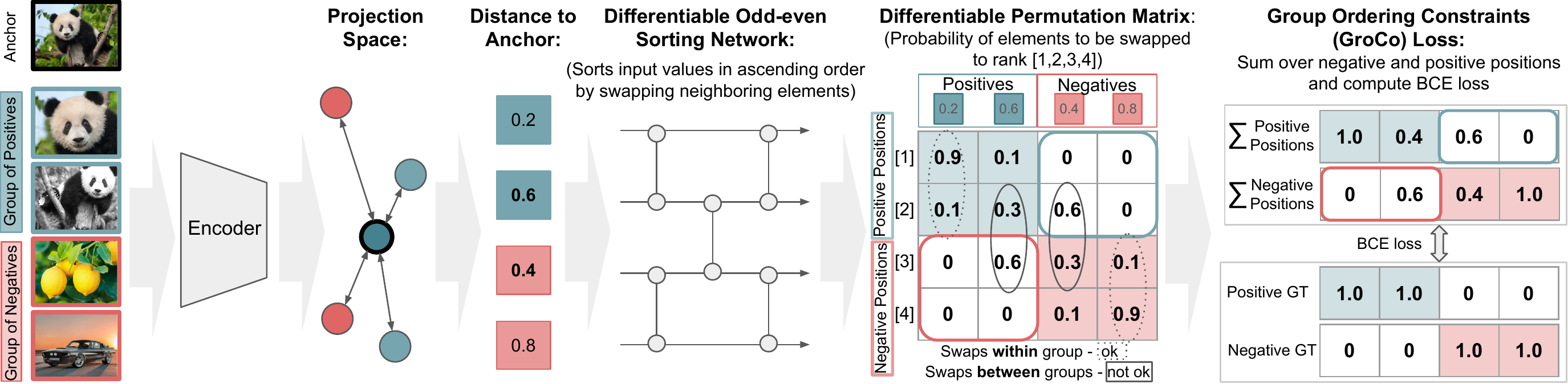}%
    \vspace{-1mm}
    \caption{
    Overview of the proposed loss: Distances of positives and negatives are computed with respect to an anchor.
    The concatenated distances are sorted via a differentiable sorting network that computes the \textit{swapping probability}. 
    The result is a differentiable permutation matrix, in which the column values can be considered as the probabilities of sorting the elements to the corresponding positions. 
    To enforce only the relationships \textit{between groups}, we sum over the positive and negative rows of the permutation matrix. 
    The loss is then computed as the BCE between the row-wise entries and the ground truth.
    \label{fig:groco_overview}}
    \vspace{-0.25cm}
\end{figure*}
However, it has been noted that the idea of a pairwise contrastive loss also has some limitations, such as the alignment of the embedding space based on individual pairs.
Several attempts have been made to address this issue, e.g., combining the contrastive idea with concepts based on local neighborhoods, such as clustering (SwAV~\cite{caron2020unsupervised}), or minimizing distances between multiple positive pairs for the same instance together (Whitening~\cite{ermolov2021whitening}).
Another limitation of the contrastive loss is that 
is that the embedding space is optimized with respect to all negatives, i.e., even negatives that are far away from the anchor will contribute to the optimization of the representation. 
Other methods were proposed to address this issue, such as hard negative selection by controlling the hardness of examples~\cite{robinson2020contrastive} or negative selection by sparse support vectors~\cite{shah2022max}. 
Nevertheless, these methods still require manual selection of the hardness level~\cite{robinson2020contrastive} or incur an additional optimization cost~\cite{shah2022max}.

To shift away from the concept of minimizing resp.~maximizing all pairwise distances, this paper proposes a variation of the contrastive learning formulation, namely Group Ordering Constraints (GroCo). 
The idea of GroCo is that positive and negative distances should be sorted in a way that any positive should be closer to an anchor image than any negative, thus forming a group of positive pairs and a group of negative pairs. 
The idea is illustrated in Figure~\ref{fig:idea}. 
In comparison to pairwise contrastive losses, the GroCo loss combines the distance information of groups of positive and negative pairs and optimization mainly depends on incorrectly ``sorted'' pairs.
To enforce the group ordering constraints in the projection space, we propose the idea of \textit{learning by sorting}:
we suggest sorting positives and negatives by distance to the anchor image in a differentiable way and swapping them if they are in the wrong order. 
This leads to a more holistic approach considering all relationships between data points, thereby better utilizing and optimizing the embeddings (esp.~for multiple positive pairs), and leading to improved down-stream performance.
To create an end-to-end training pipeline, we leverage recent advances in differentiable sorting~\cite{cuturi2019differentiable,grover2019stochastic,petersen2022thesis,petersen2021diffsort,petersen2021learning,petersen2022monotonic,petersen2022differentiable}.
Specifically, we utilize a differentiable sorting algorithm to obtain a differentiable permutation matrix for sorting a list of distances to the positive and negative images, as shown in Figure~\ref{fig:groco_overview}. 
If we would know the full ground truth orderings among positives and negatives (such as which positive sample should be closer to the anchor than another positive sample), we could create a ground truth permutation matrix, and calculate how much the predicted permutation matrix would deviate from the ground truth one~\cite{grover2019stochastic,cuturi2019differentiable,petersen2021diffsort,petersen2022monotonic}. 
Because we do not know the ground truth distance ordering within the positive or the negative groups, we propose the GroCo loss as a relaxed formulation of the original sorting supervision that captures how many negative elements appear in the positive positions and vice versa. 
The proposed GroCo loss alleviates some aspects of vanilla contrastive learning: first, it treats positive and negative pairs as groups instead of individual pairs, and second, the resulting group ordering focuses on \textit{optimizing the local neighborhood} around an anchor image by mainly optimizing too close negative and too distant positives, rather than optimizing all data points at once.
Thus, it implicitly also focuses on the strongest positive (furthest from the anchor) and strongest negative (closest to the anchor) examples.

To show the capabilities of the proposed approach, we evaluated
it on various competitive self-supervised learning benchmarks, namely in the context of linear probing, $k$-NN classification, transfer learning, as well as image retrieval.
The evaluation shows that the model trained via group ordering constraints outperforms contrastive learning frameworks in linear probing and transfer learning and excels in the context of shaping local neighborhoods on the tasks such as $k$-NN classification and image retrieval. 

\noindent The contributions of this work are summarized as follows:
\begin{itemize}[leftmargin=1.em,itemsep=-1pt,topsep=3pt]

    \item We advance the concept of contrastive learning by introducing Group Ordering Constraints (GroCo) that treat positive and negative elements as groups rather than individual pairs as in conventional contrastive learning.
    
     \item {To derive a loss that optimizes the proposed constraints, we harness recent differentiable sorting methods and obtain a loss that suggests sorting positive and negative elements and swapping them if they are in the wrong order --- thus, we introduce a new contrastive learning method called \textit{learning by sorting}.}

    \item The proposed method provides embeddings that achieve competitive performance in linear probing and are especially suitable to model the local neighborhoods and outperform 
    contrastive learning frameworks on a wide range of nearest-neighbor tasks.   
\end{itemize}

\vspace{-.5em}
\section{Related Work}
\vspace{-.25em}
\subsection{Self-supervised Representation Learning}
\vspace{-.3em}

\noindent \textbf{Contrastive methods:} Over the last years, self-supervised learning methods that enforce the model to be robust to different image distortions achieved great performance improvements in self-supervised learning~\cite{chen2020simple, he2020moco,chen2020big,chen2020improved}. 
Such methods generally rely on sampling two augmented views of the image---a positive pair---and minimize the distance between those in the embedding space. To prevent the model from learning a trivial solution for any input, contrastive methods introduce the concept of a negative pair, i.e., two different images, to \textit{contrast} positive against negative pairs.
While earlier contrastive methods relied on the triplet loss~\cite{schroff2015facenet}, the probably most prominent method in many self-supervised learning scenarios is the InfoNCE loss, which is often referred to as a contrastive loss~\cite{chen2020simple,he2020moco,radford2021learning,akbari2021vatt}, which requires accumulating strong negatives via a memory bank~\cite{he2020moco} or a large batch size~\cite{chen2020simple}. 
Many extensions have been proposed to further improve the performance of this idea: data augmentation strategies~\cite{chen2020simple,xiao2020should}, projection head design~\cite{chen2020big}, hard negative sampling~\cite{robinson2020contrastive}, increasing the richness of positives with nearest neighbours~\cite{dwibedi2021little}, or mitigating the effect of false negatives~\cite{huynh2022boosting}. 
The variation of the contrastive method proposed in this work should not be considered as opposed but rather as orthogonal to other approaches relying on positive and negative pairs because it changes the loss function itself and can therefore be used, e.g., on top of other techniques.

\vspace{.23em}
\noindent \textbf{Alternative methods:} There are also methods~\cite{chen2021exploring,grill2020bootstrap} that do not rely on negatives and only maximize agreement between positive views. 
Such methods prevent collapsing of the representation space by using asymmetric architectures applied to different views~\cite{chen2021exploring,grill2020bootstrap}, an additional teacher network~\cite{grill2020bootstrap,caron2021emerging}, stop gradient~\cite{chen2021exploring,grill2020bootstrap,caron2021emerging}, feature whitening~\cite{ermolov2021whitening}, or information maximization~\cite{zbontar2021barlow,bardes2021vicreg}.
Another set of methods~\cite{caron2020unsupervised,caron2018deep,asano2019self} utilizes clustering of latent embeddings. 
ReSSL~\cite{zheng2021ressl} leverages relations in an embedding space in a self-labeling way, namely aligning the similarities between weakly-augmented images to the similarities to the similarities between strongly augmented images. 
SwAV~\cite{caron2020unsupervised} additionally proposes sampling more augmentations in a multi-crop setting, where two full-size augmented images are sampled together with several smaller crops, and Whitening~\cite{ermolov2021whitening} utilized more full-resolution samples. 
While currently methods relying on positive samples seem to outperform their classical contrastive counterparts, the we show that especially local neighborhood learning can profit from relying on positive and negative samples.

\vspace{-0.7mm}
\subsection{Differentiable Sorting and Ranking}
\vspace{-0.7mm}

Differentiable sorting and ranking methods provide a pipeline that allows training neural networks with ordering supervision in an end-to-end fashion with gradient descent~\cite{
Blondel2020-FastSorting,cuturi2019differentiable,grover2019stochastic,petersen2022thesis,petersen2021diffsort,petersen2021learning,petersen2022monotonic,petersen2022differentiable,prillo2020softsort}. 
{Earlier pairwise learning-to-rank methods, such as RankNet~\cite{burges2005learning} or LambdaRank~\cite{burges2006learning}, and listwise methods, such as SoftRank~\cite{taylor2008softrank} or ListNet~\cite{cao2007learning},  are mostly based on heuristics and aim to optimize ranking metrics, e.g., NDCG. 
Many of the latest differentiable sorting approaches~\cite{
cuturi2019differentiable,grover2019stochastic,petersen2022thesis,petersen2021diffsort,petersen2022monotonic,petersen2022differentiable,prillo2020softsort} focus on obtaining a differentiable relaxation for the sorting operator.}
The sorting operator can be seen as a function returning a permutation matrix that indicates the permutation necessary to sort the sequence of values (the matrix that multiplied with an input vector returns a sorted output vector.)
In this context, differentiable sorting refers to relaxing the (hard) permutation matrix to a differentiable permutation matrix via continuous relaxations.
The differentiable permutation matrix for a given sequence of values, which can, e.g., be scores predicted by a neural network, can then be used to compute the loss by comparison to a ground truth permutation matrix.
Recently, multiple methods for relaxing the permutation matrix have been proposed, including an argsort approximation by unimodal row-stochastic matrices~\cite{grover2019stochastic,prillo2020softsort}, a formulation of entropy-regularized optimal transport~\cite{cuturi2019differentiable}, as well as networks of differentiable swap operations (differentiable sorting networks)~\cite{petersen2021diffsort,petersen2022monotonic}. 
The latter method composes the full permutation matrix as a product of permutation matrices that arise from comparing only two elements at a time (usually neighbors) and either swapping them or not swapping them.
Practically, differentiable sorting has been leveraged in various contexts, including recommender systems~\cite{lee2021differentiable,swezey2021pirank}, image patch selection~\cite{cordonnier2021differentiable}, selection experts in multi-task learning~\cite{hazimeh2021dselect}, attention mechanisms~\cite{zhan2021bi}, and audio representation learning~\cite{Carr21Audio}. 
To the best of our knowledge, the proposed method is the first work to leverage ordering supervision for self-supervised learning of visual representations.

\vspace{-.25em}
\section{Method}
\vspace{-.25em}

Given a dataset of images $\{x_i\}_{i=1}^M \subseteq \mathcal{X}$, the goal is to learn an encoder $g: \mathcal{X} \to \mathbb{R}^d$ that extracts image representations that can later be used for downstream tasks.

\vspace{-.1em}
\subsection{Training Pipeline}
\vspace{-.1em}

As in standard contrastive losses, the proposed method considers several augmented views of the same image as positive examples, which should be close together in the embedding space, and different images as negative examples, which should be apart in the embedding space. 
Starting from mini-batches of $B$ images, $m \ge 2$ randomly augmented views are generated for each image, resulting in $m \cdot B$ data points overall per batch. Note that if $m=2$, the proposed method is close to the original contrastive learning setup~\cite{huynh2022boosting,chen2020simple,chen2021exploring,bardes2021vicreg,zbontar2021barlow}. 
The augmented views are processed with the encoder network $g(\cdot)$ and an MLP projection head $h(\cdot)$, that maps images to the latent space where distances between views are calculated. 
For each data point serving as an anchor $x^a$, there are $m - 1$ positive examples $\{x^p_i\}_{i=1}^{m-1}$  
and $m\cdot(B - 1)$ negative examples $\{x^n_i\}_{i=1}^{m\cdot(B - 1)}$. 
The measure of distance between data points is the cosine distance defined as: $\mathsf{d}(x, y) = - \frac{x^\top y}{\left\lVert x\right\rVert \left\lVert. y\right\rVert}$.

\vspace{-.1em}
\subsection{Group Ordering Constraints (GroCo)}
\vspace{-.1em}

In order to consider positives and negatives not individually but instead as a group,
the proposed loss extends the contrastive loss %
to the idea that \textit{the group of positives} should be closer to the anchor image than \textit{the group of negatives} in the embedding space resulting in \textit{group ordering constraints (GroCo)}. 
To simplify the notation, the distance between data point $x^a$ and its positive $x^p_i$ and negative examples $x^n_i$ is denoted as $d^p_i = \mathsf{d}(x^a, x^p_i)$ and $d^n_i = \mathsf{d}(x^a, x^n_i)$. 
Assuming that $K$ positives $x^p_1,...,x^p_{K}$  are ordered with respect to their distances to the anchor $x^a$ as 
$d_1^p \le ... \le d_{K}^p$
and $N$ negatives 
$x^n_1,...,x^n_{N}$ 
as 
$d_1^n \le ... \le d_{N}^n$,
then the \textit{group ordering constraints} can be defined as 
\begin{equation}
\label{eq:groco}
    d_1^p \le ... \le d_{K}^p~\mathlarger{\mathlarger{\pmb{<}}}~d_1^n \le ... \le d_{N}^n.
\end{equation}

We note that all elements are considered in the constraints and the relevant constraint is the (bold) $<$ in the center.
{We remark that, although the constraint is already fulfilled if the largest positive distance $d_{K}^p$ is smaller than the smallest negative distance $d_1^n$, it is suboptimal to define loss only on those elements.
Comparing only the smallest negative and largest positive ignores other negatives (e.g., the second smallest) and positives (e.g., the second largest) that might also be misaligned, and such a loss would ignore them. 
In the next section, we propose our novel loss that optimizes the GroCo constraints implicitly.
}

\vspace{-.1em}
\subsection{Learning by Sorting}
\vspace{-.1em}

To enforce the constraint, the GroCo loss leverages recent advances in differentiable sorting~\cite{petersen2021diffsort,petersen2022monotonic}, which allow to derive a loss that fulfills ordering constraints. Namely, the training procedure can be seen as sorting positives and negatives in the embedding space with respect to an anchor image and swapping them if they are in the incorrect order, which relates to the proposed idea of \textit{learning by sorting}.

\begin{figure}[]
    \centering 
    \begin{subfigure}{\linewidth}
         \centering 
        \includegraphics[width=0.95\textwidth]{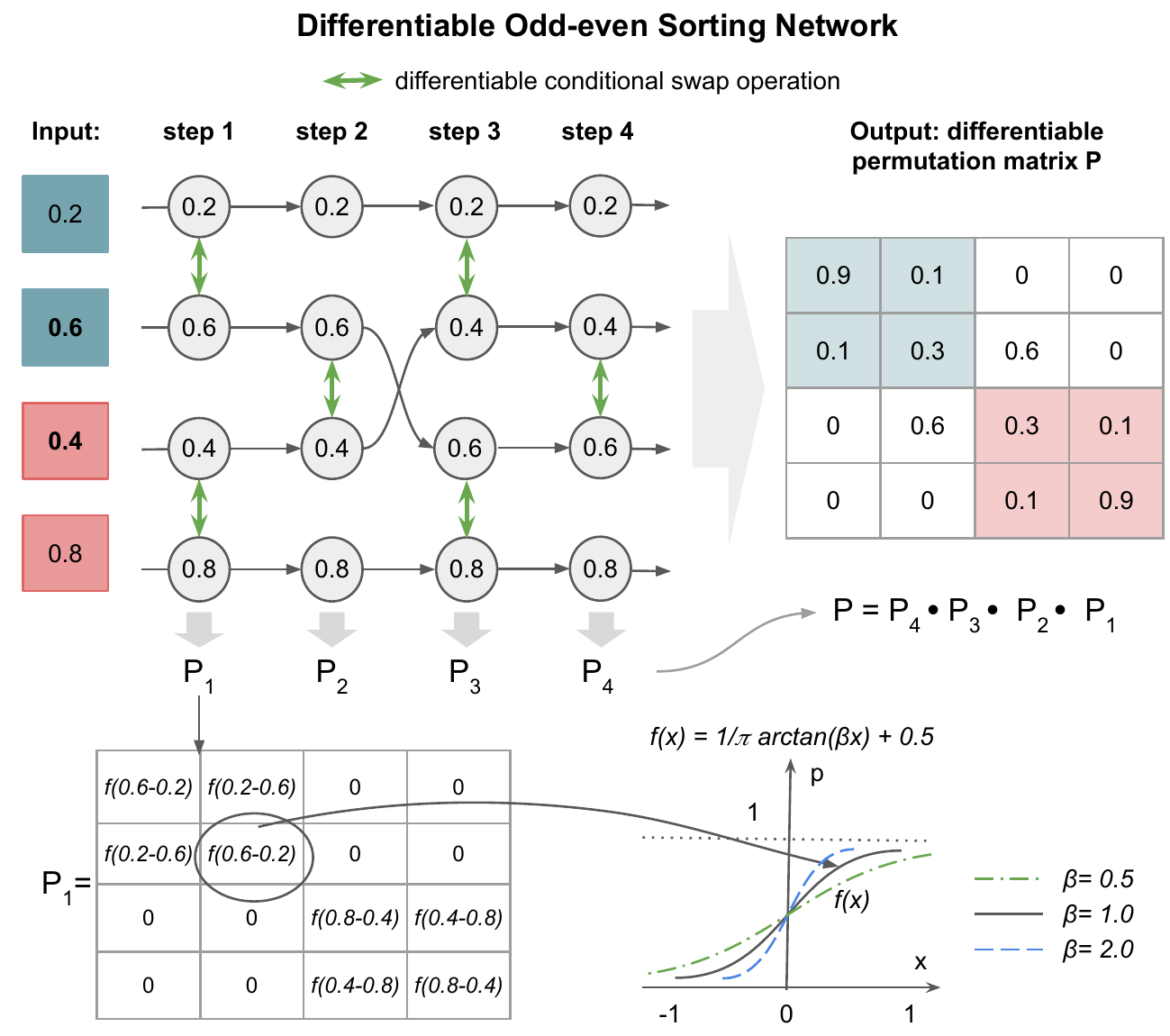}
    \end{subfigure}
    \caption{
    Overview of a differentiable sorting network with odd-even sorting: 
    The network compares neighboring elements starting from odd and even indices alternatingly in each step and applies a differentiable swap operation if elements are in the wrong order. The swap operations on each step~$s$ also define a differentiable permutation matrix $P_s$. The network output is a differentiable permutation matrix $P$, defined as the multiplication of matrices of each step.
    \label{fig:sorting}}
    \vspace{-5mm}
\end{figure}

\vspace{-3mm}
\subsubsection{Differentiable Sorting Networks}
\label{sec:diff-sort-net}
\vspace{-1mm}

This section provides a review of the differentiable sorting algorithm \textit{differentiable sorting networks}~\cite{petersen2021diffsort} used for the proposed loss function. 
Note that ``networks'' in ``sorting networks'' are not ``neural networks'' but instead refers to a category of sorting algorithms in the computer science literature~\cite{knuth1998sorting} with no trainable parameters. %

Differentiable sorting networks, e.g., based on the odd-even sorting network, sort an input sequence of $K + N$ elements in non-descending order as shown in Figure~\ref{fig:sorting}. 
They are defined as the concatenation of functions, e.g., representing the swap operations in each layer of an odd-even sorting network, where each function refers to one step of sorting, and pairs of elements of the input sequence are compared and swapped if they are in the wrong order via a \textit{conditional swap operation}. 
For the odd-even sorting network, the algorithm compares neighbored elements on odd and even indices alternatingly and requires $K + N$ steps to sort a given input sequence of length $K + N$. 
By relaxing the conditional swap operator to a differentiable one, sorting networks can be made differentiable~\cite{petersen2021diffsort}. The conditional swap operation for elements $(d_i, d_j)$ where $i < j$ can be defined as $d'_i = \min(d_i, d_j)$,  $d'_j = \max(d_i, d_j)$, and the differentiable relaxation~\cite{petersen2022monotonic} of this operation is: 
\begin{equation}
\begin{split}
\label{eq:softswap}
\kern-2.4emd'_i = \mathrm{softmin}(d_i, d_j) &= d_i f(d_j - d_i) + d_j f(d_i - d_j),\kern-1em \\
\kern-2.4em d'_j = \mathrm{softmax}(d_i, d_j) &=  d_i f(d_i - d_j) + d_j f(d_j - d_i)\kern-1em \\
\end{split}
\end{equation}
with\vspace{-.75em}
\begin{equation}
\label{eq:swap}
f(x) = \frac{1}{\pi} \arctan(\beta x) + 0.5 .
\end{equation}
The hyperparameter $\beta > 0$ denotes an inverse temperature. For $\beta\to\infty$, the relaxation converges to the discrete swap operation.
The differentiable conditional swap operation for the elements $(d_i, d_j)$ can be defined as a permutation matrix $P_{swap(d_i, d_j)} \in R^{(K+N)\times (K+N)}$, which is an identity matrix except for entries $P_{ii}, P_{ij}, P_{jj}, P_{ji}$ defined as: 
\begin{equation}
\begin{split}
\begin{aligned}
    &P_{ii} = P_{jj} = f(d_j - d_i), \\
    &P_{ij} = P_{ji} = f(d_i - d_j). \\
\end{aligned}
\end{split}
\end{equation}
The permutation matrix $P_s$ for step $s$ is the product of matrices corresponding to independent (and thus parallel) swap operations in this step $P_s= \prod_{i\in R}{P_{swap(d_i, d_{i+1})}} $, where $R$ is the set of odd indices if $s$ is odd and the set of even indices if $s$ is even.  
The complete permutation matrix $P$ is defined as $P = P_{K+N}\cdot ... \cdot P_1$. 
In the discrete case, each column of a permutation matrix has exactly one entry of $1$, indicating the position where the element that corresponds to this column should be placed. 
In the relaxed version, column values can be seen as a distribution over possible positions of the element. 
If the correct order of input values is known, we can create a ground truth matrix $Q$ and define the loss as 
$ ~ L = \frac{1}{(K+N)^2} \sum_{i,j} \operatorname{BCE}(P_{ij}, Q_{ij})~$ where $\operatorname{BCE}$ refers to binary cross-entropy.
\vspace{-.9em}
\subsubsection{GroCo Loss}
\vspace{-.35em}
\label{sec:groco_loss}
If there was a known ground truth order of positives and negatives, the loss could be calculated directly based on this ground truth permutation matrix.
However, as the algorithm is based on random augmentations, there are no known orders among positives and among negatives. Thus, the only information available is whether a pair belongs to the positive or the negative group.

To derive a loss to fulfil this group ordering constraint, we start by ordering positives and negatives first separately with respect to the computed distances to the anchor image as $d_1^p \le ... \le d_{K}^p$ for positives and $d_1^n \le ... \le d_{N}^n$ for negatives.
As shown in Figure~\ref{fig:sorting}, positive and negative distances are then concatenated in a list as:
\vspace{-.25em}
\begin{equation}
    [d_1^p,...,d_{K}^p, d_1^n,...,d_{N}^n].
\end{equation}
Even though elements in the positive and negative are ordered, it is still open if the constraint $d_i^p < d_j^n$ is fulfilled for any $1 \le i \le K$ and $1 \le j \le N$.

A differentiable sorting network is applied to the concatenated list and a differentiable permutation matrix is obtained for sorting the list in non-descending order.
As shown in~\ref{fig:sorting}, values in the permutation matrix column can be seen as probabilities to sort the corresponding element to the different positions, e.g., $P_{11}$ would be the probability for assigning the first element in the list ($d_1^p$) to position $1$, $P_{21}$ to position $2$, etc. 
Therefore, the sum of the first $K$ elements in a column can be considered as a probability being sorted inside the first $K$ elements. 
Thus, for a permutation matrix of size $(K+N)\times(K+N)$ the sum of the first $K$ rows results in probabilities of being sorted in \textit{positive places} and the later columns (from  $K + 1$ to $K + N$) in \textit{negative places}.
To enforce positives to be sorted in the positive places and negatives in the negatives places, the respective loss (with $\mathbbm{1}$ as the indicator function) is defined as 
\vspace{-.5em}
\begin{align}
\label{eq:loss}
   L= \frac{1}{2(K + N)} & \sum_{i=1}^{K + N}  \bigg( \operatorname{BCE}\Big({\textstyle\sum_{k=1}^{K}}{P_{ki}}, \mathbbm{1}_{i \le K}\Big) ~+ \\
   & ~~~~~~~~\,+ \operatorname{BCE}\Big({\textstyle\sum_{k=K+1}^{K + N}{P_{ki}}}, \mathbbm{1}_{i > K}\Big) \bigg)\notag\,.
\end{align}
As illustrated in Figure~\ref{fig:groco_overview}, the proposed loss is a relaxation of the sorting supervision in a way that it considers two types of swap operations: swap operations \textit{within} the group of positive and negative samples, which should not contribute to the loss, and swap operations \textit{between} the groups, which violate the positive-negative ordering assumption and which are used as the optimization criterion.

\myparagraph{Role of $\beta$.}
One relevant hyper parameter is the inverse temperature $\beta$ in differentiable swap operation (Equation~\ref{eq:swap}), which corresponds to the degree of relaxation of the swap operation that converges to a discrete case when $\beta\to\infty$ (Figure~\ref{fig:sorting}). 
Therefore with lower $\beta$ the swap operation is more ``soft'', which is beneficial for optimization, but the relaxation error accumulated by each step is larger, and vice versa in the case of larger $\beta$.
With higher $\beta$ even a small difference between values results in a high probability for a swap or not swap operation, resulting in a smaller margin between the positive and negatives group. 

\myparagraph{Number of samples.}
Since the strongest negatives have the strongest effect on the loss function, the selection can be limited to only the top-$N$ strongest negatives. Further, as more negatives also result in more layers in the sorting network and each layer contributes to the overall differential permutation matrix, more layers also result in a softer swap probabilities. Therefore $\beta$ should be selected based on a number of elements to sort. An ablation study on this effect is given in Section~\ref{sec:ablation}. 

\myparagraph{Role of Pre-ordering.}
Practically, positives and negatives are pre-ordered among themselves before concatenating and forwarding them to the differentiable sorting network. 
While the loss will still contrast positives to negatives no matter if the input is ordered or not, it shows that pre-ordering improves the overall performance of the GRoCo loss. 
This can be attributed to the fact that sorting networks perform comparisons between neighboring elements and swap them if they are in the wrong order. 
If the input was not pre-ordered and, thus, distances from positive and negative pairs would be mixed, this would result in additional swap operations.
Using pre-ordered inputs the focus lies on comparing the strongest positives with the strongest negatives. 
In this way, elements are considered as a group, and the borders of the groups or their overlapping parts are emphasized in the loss.
In Section~\ref{sec:ablation} and Figure~\ref{fig:loss_optimization}, we provide additional discussions and illustrate this behavior.

\vspace{-.1em}
\section{Experimental Evaluation}
\vspace{-.2em}
\subsection{Implementation Details}
\vspace{-.2em}
\label{sec:impl_details}

Unless stated otherwise, the following setup is used for all experiments:

\myparagraph{Model.} Following previous works~\cite{chen2020simple,grill2020bootstrap,caron2020unsupervised,chen2021exploring} Resnet50~\cite{he2016deep} is used as the encoder $g(\cdot)$ and an MLP block consisting of three fully connected layers with a size of 2048 and followed by a batch normalization layer~\cite{ioffe2015batch} is used as the projection head $h(\cdot)$.
All batch normalization layers except the last one are followed by a ReLU activation. 
The dimensionalities of the representation space and the latent space are both $2\,048$ as in~\cite{chen2021exploring}.

\myparagraph{Training.} 
Following previous works~\cite{chen2020simple,grill2020bootstrap,caron2020unsupervised,chen2021exploring}, we use the train set of the ImageNet ILSVRC-2012 dataset~\cite{russakovsky2015imagenet} for self-supervised training without any human annotation. 
To create $m$ augmented views per image (considering $m=2,3,4$), the DINO augmentation setup~\cite{caron2021emerging} is used.
The model is trained with the SGD optimizer~\cite{you2017large} with a learning rate of $6.0 \times (\text{batch size} / 256)$ for 100 epochs and $3.0 \times (\text{batch size} / 256)$ for 200 and 400 epochs. We use a cosine scheduler without restarts~\cite{loshchilov2016sgdr} and $10$ epochs warm-up for 200 and 400 epochs training and $1$ epoch linear warm-up for 100 epochs training (we find it beneficial for our method but not for SimCLR).
During training, the stop gradient operation is used following the self-supervised learning setups of~\cite{chen2020simple,caron2021emerging,grill2020bootstrap}; specifically, stop gradient is performed during distance computation $\mathsf{d}(x^a, x^\cdot_i) = \mathsf{d}(x^a, \texttt{stop\_grad}(x^\cdot_i))$. 
While training with stop gradient does not show a direct impact on the overall performance, we observed that it allows training with larger variations of hyperparameters while maintaining stable performance. 
By default, the top $N=10$ strongest negatives are sampled from the batch and an inverse temperature of $\beta=1$ is used.
Due to resource constraints, the model is trained with a batch size of $1\,024$ with mixed precision. 
On an 8-GPU (NVIDIA A6000) server, training for $100$ epochs with $m=2$ views takes approximately $22$ hours.

\begin{table}[]
    \centering
    \resizebox{\columnwidth}{!}{
		 \begin{tabular}[t]{lcc|ccc}
			    \toprule 
			     Method & BS & Views &  100 ep & 200 ep & 400 ep  \\
			     \midrule
                   \multicolumn{6}{c}{Linear Probing (Top-1)}\\
			     \midrule
                  Max-Margin~\cite{shah2022max} & 256 &2$\times$224 & 63.8 & - & - \\
			     MoCo v2$\dagger$~\cite{chen2020improved}~\cite{chen2021exploring}& 256 & 2$\times$224 & 67.4 & 69.9 & 71.0 \\
                 SimCLR$\dagger$~\cite{chen2020simple}~\cite{chen2021exploring}& 4096 & 2$\times$224 & 66.5 & 68.3 & 69.8 \\
                 \rowcolor{almond} GroCo (ours) & 1024 & 2$\times$224 & 69.2 & 70.4 & 71.1 \\
			     \rowcolor{almond} GroCo (ours) & 1024 & 4$\times$224 & \textbf{69.6} & \textbf{70.6} & 71.3 \\
                \midrule
			     SimSiam~\cite{chen2021exploring} & 256 &2$\times$224 & 68.1 & 70.0 & 70.8 \\
                  VICReg~\cite{bardes2021vicreg} & 2048 & 2$\times$224 & 68.6 & - & -  \\
                  Barlow Twins~\cite{zbontar2021barlow} & 2048 & 2$\times$224 & 68.7 & - & - \\
                  SwAV$\dagger$~\cite{caron2020unsupervised}  & 4096 & 2$\times$224 & 66.5 & 69.1 & 70.7 \\
                  ReSSL~\cite{zheng2021ressl} & 256 & 2$\times$224 & - & 69.9 & - \\
			     BYOL$\dagger$~\cite{grill2020bootstrap} & 4096 & 2$\times$224 & 66.5 & \textbf{70.6} & \textbf{73.2}\\
                  Whitening~\cite{ermolov2021whitening} & 4096 & 4$\times$224 & \underline{69.4} & - & \underline{72.6} \\

                 \midrule
                \multicolumn{6}{c}{$k$-NN (weighted, $k$=20)}\\
                \midrule

                  MoCo v2~\cite{chen2020improved} & 256 & 2$\times$224 & - & 55.6 & - \\
                  SimCLR$\dagger$~\cite{chen2020simple}~\cite{chen2021exploring} & 4096 & 2$\times$224 & 53.8 & 57.2 & 59.2\\
                   \rowcolor{almond} GroCo (ours) & 1024 & 2$\times$224 & \underline{60.5} & \underline{62.9} & \underline{64.0} \\
                  \rowcolor{almond} GroCo (ours) & 1024 & 4$\times$224 & \textbf{61.8} & \textbf{63.6} & \textbf{64.8} \\
                  \midrule
			     SimSiam~\cite{chen2021exploring} & 256 & 2$\times$224 & 57.4 & - & - \\
                 SwAV~\cite{caron2020unsupervised}  & 4096 & 2$\times$224 & - & - & 61.3 \\
        
			\bottomrule
		\end{tabular}
		}
  \vspace{-1mm}
  \caption{\textbf{Comparison to state-of-the-art in linear probing $k$-NN classification on ImageNet.}~We report results for training for $100, 200, 400$ epochs. \textit{Backbone=Resnet50.} $\dagger$~denotes improved reproductions from SimSiam~\cite{chen2021exploring}. \label{tab:sota_linear}}
  \vspace{-2mm}
\end{table}

\vspace{-.3em}
\subsection{Evaluation Procedure}
\vspace{-.7em}
\myparagraph{Linear Probing.} 
Linear probing allows to evaluate the learned embedding space by linear evaluation~\cite{chen2020simple,grill2020bootstrap,caron2020unsupervised,chen2021exploring}, capturing the linear separability of classes. For this, a linear classifier is trained on frozen representations in a fully-supervised way using the ImageNet train set.  
We follow the standard protocol~\cite{chen2021exploring} to train the linear classifier.

\myparagraph{$k$-NN Evaluation.}
To analyze the local properties of the learned representation, namely how often neighbored data points correspond to the same semantic class, we further evaluate with respect to nearest neighbor classification, predicting the class by a simple weighted $k$ nearest neighbor classifier ($k$-NN) with $k=\{1,10,20\}$ based on cosine distance as used in~\cite{caron2021emerging,caron2020unsupervised}.
Again, we use the ImageNet train set for supervision and test on the val set.

\subsection{Comparison to State-of-the-Art}

\begin{table}[]
    \centering
    
    \resizebox{\columnwidth}{!}{
		 \begin{tabular}[t]{lccc|cccc}
			    \toprule 
			     \multirow{2}{*}{Method} & \multirow{2}{*}{Epochs} & Batch & \multirow{2}{*}{Views} & \multicolumn{2}{c}{Oxford} & \multicolumn{2}{c}{Paris}\\
			      & &  Size & & M & H & M & H \\
			     \midrule
			     SimSiam~\cite{chen2021exploring} & 100 & 256 & 2$\times$224 & 26.89 & 7.04 & 46.92 & 19.31 \\
			     MoCo v2~\cite{chen2020improved} & 200 & 256 & 2$\times$224 & 23.28 & 5.07 & 42.8 & 17.33 \\
                  SimCLR~\cite{chen2020simple} & 400 & 4096 & 2$\times$224 & 23.27 & 4.56 & \underline{46.93} & \underline{20.19} \\
                  SwAV~\cite{caron2020unsupervised} & 400  & 4096 & 2$\times$224 & \underline{28.01} & \textbf{8.35} & 46.23 & 17.4 \\
                  \rowcolor{almond} GroCo (ours) & 400 & 1024 & 2$\times$224 & \textbf{29.37} &  \underline{7.11} & \textbf{54.95} & \textbf{26.26} \\

			\bottomrule
		\end{tabular}
		}
  \vspace{-1mm}
  \caption{
  \textbf{Comparison to state-of-the-art in image retrieval.} We evaluate image retrieval performance on the Medium (M) and Hard (H) splits of the revisited Oxford and Paris datasets~\cite{radenovic2018revisiting}.
  We evaluate nearest neighbor retrieval performance with ImageNet-trained encoders and report MAP. \textit{Backbone=Resnet50.} 
  \label{tab:sota_retrieval}}
  \vspace{-4mm}
\end{table}

\begin{table*}[]
    \centering
    \resizebox{\linewidth}{!}{
		 \begin{tabular}[t]{lcc|ccccc ccccc c|c}
			    \toprule 
Method & Epochs & BS & Aircraft & Caltech101 & Cars & Cifar10 & Cifar100 & DTD & Flowers & Food & Pets & SUN397 & VOC2007 & Average \\ 
\midrule
MoCo v2 & 200 &  256 & 20.1 & 78.9 & 12.3 & 86.9 & 61.4 & 68.7 & 67.3 & 47.2 & 67.2 & 46.9 & 73.9 & 57.35 \\
SimCLR  & 400 & 4096 & 19.3 & 77.7 & 15.3 & 85.4 & 61.0 & 70.2 & 72.5 & 51.2 & 68.2 & 49.0 & 73.1  & 58.45 \\
SimSiam & 100 & 256 & 25.4 & 80.1 & 17.3 & \underline{87.4} & {65.5} & 69.3 & \underline{77.3} &  \underline{52.5} & \underline{72.5} & 50.0 & 72.8 & 60.92\\

SwAV & 400 &  4096 & \underline{25.8} & \textbf{82.2} & \underline{17.8} & \textbf{88.6} & \textbf{66.0} & \textbf{71.1} & 74.8 & 50.3 & 70.4 & \textbf{52.9} & \textbf{75.6} & \underline{61.41} \\

\rowcolor{almond} GroCo (ours) & 400 & 1024 & \textbf{27.6} & \underline{81.2} & \textbf{19.1} & 86.8  & \underline{65.8} & \underline{71.0} & \textbf{79.2} &  \textbf{56.3} & \textbf{80.6} & \underline{52.3} & \underline{74.6} & \textbf{63.14} \\

\bottomrule
		\end{tabular}
		}
  \caption{\textbf{Comparison to state-of-the-art transfer performance in k-NN classification on 11 classification datasets.} Models are pre-trained on ImageNet.
  \textit{Backbone=Resnet50, views=2x224.} \label{tab:simclr_transfer}}
  \vspace{-4mm}
\end{table*}

\begin{table}[]
    \centering
    \resizebox{\columnwidth}{!}{
		 \begin{tabular}[t]{lcc|ccc}
			    \toprule 
			     Method & BS & Views &  100 ep & 200 ep & 400 ep  \\
			     \midrule
                   \multicolumn{6}{c}{Linear Probing (Top-1)}\\
                   \midrule
                SwAV~\cite{caron2020unsupervised}  & 4096 & 2$\times$224 + 6$\times$96 & 72.1 & 73.9 & 74.6 \\
                \rowcolor{almond} GroCo (ours) & 1024 & 2$\times$224 + 6$\times$96 & 71.8 & 72.9 & 73.7 \\
                \midrule
                \multicolumn{6}{c}{$k$-NN (weighted, $k$=20)}\\
                \midrule
                SwAV~\cite{caron2020unsupervised} & 4096 & 2$\times$224 + 6$\times$96 & 61.7 & 63.7 & 64.9 \\		    
			     \rowcolor{almond} GroCo (ours) & 1024 & 2$\times$224 + 6$\times$96 & 62.3 & 64.2  & 65.2 \\

			\bottomrule
		\end{tabular}
		}
  \caption{\textbf{Comparison to SwAV for the multi-crop augmentation strategy.}  We report results for training for $100, 200,$ and $400$ epochs.  \textit{Backbone=Resnet50.} \label{tab:sota_multicrop}}
  \vspace{-4mm}
\end{table}

We start with a comparison of the proposed method to state-of-the-art self-supervised learning methods in linear probing and $k$-NN evaluation on the ImageNet~\cite{russakovsky2015imagenet}, and in image retrieval on the revised Oxford and Paris dataset~\cite{radenovic2018revisiting}, as well as in transfer learning. 

\myparagraph{Linear Probing.} 
In the case of linear probing (Table~\ref{tab:sota_linear}), the proposed method is compared to contrastive baselines using positive and negative samples, namely Max-Margin~\cite{shah2022max}, SimCLR~\cite{chen2020simple}, and MoCo v2~\cite{chen2020improved}, as well as to alternative methods.
We observe that, in the given setting, the proposed loss is able to improve above all contrastive baselines and even outperforms most strong alternative baselines except BYOL~\cite{grill2020bootstrap} and Whitening~\cite{ermolov2021whitening} (in 400 epochs setup) that use $\times4$ larger batch size and/or an additional teacher network. Another finding is that in the context of linear probing, having four positive samples does not significantly improve the approach compared to only two samples. We attribute this to the fact that missing fine-grained differences in the local neighborhood are compensated by the linear layer training, thus the initial pretraining is less relevant with respect to the local neighborhood in this setting.

\myparagraph{$k$-NN Evaluation.} 
Further, the proposed method is evaluated with respect to the $k$-NN performance and compared to state-of-the-art methods that officially released weights (Table~\ref{tab:sota_linear}).
Here, it can be observed that the proposed method outperforms all methods in this setting. 
The results demonstrate that the margin, by which the loss improves over other methods, is increased compared to linear probing, where, e.g., for SwAV, the results were mainly on par ($+0.4\%$) in case of linear probing, while here they show a more substantial improvement ($+2.3\%$).
A second hint that the strength of this method is in the optimization of local neighborhoods is given by the fact that the $k$-NN setting also shows an improved performance by leveraging multiple positive examples. This can be an indication that more positive examples contribute to a better local neighborhood in this setting.

\myparagraph{Image Retrieval.} 
To further assess the potential of the proposed method in nearest neighbors-based tasks, the model is evaluated on the task of image retrieval 
in Table~\ref{tab:sota_retrieval}. 
Results are reported as the Mean Average Precision (MAP) for the Medium (M) and Hard (H) splits of the datasets as in~\cite{caron2021emerging}.
Our method outperforms all other methods in this task, confirming good local properties of learned representations.

\myparagraph{Transfer Performance.} 
Finally, we compare how well performance transfers on other datasets. 
In Table~\ref{tab:simclr_transfer}, we compare ImageNet pre-trained models in zero-shot $k$-NN evaluation on 11 classification datasets, including FGVC Aircraft~\cite{maji2013fine}, Caltech-101~\cite{fei2004learning}, Stanford Cars~\cite{krause2013collecting}, CIFAR10~\cite{krizhevsky2009learning}, CIFAR-100~\cite{krizhevsky2009learning}, DTD~\cite{cimpoi2014describing}, Oxford 102 Flowers~\cite{nilsback2008automated}, Food-101~\cite{bossard2014food}, Oxford-IIIT Pets~\cite{parkhi2012cats}, SUN397~\cite{xiao2010sun}
and Pascal VOC2007~\cite{everingham2010pascal}.  %
We observe that the proposed method improves over the SimCLR and MoCo v2 baselines in all classes and on average even outperforms the publicly available SimSiam and SwAV baselines. 
This can be particularly attributed to the improved performance on the Pets and Food datasets.
We credit the increased performance to the fact that both food and animal-related classes often appear in the ImageNet pretraining data, thus, learning a good local embedding helps with those datasets, specifically in the case of $k$-NN classification. 

\myparagraph{Multi-crop Augmentation.} 
Since computational cost grows linearly with an increasing number of augmentations, the multi-crop augmentation strategy proposed in SwAV~\cite{caron2020unsupervised} is also considered. 
The idea is to sample low-resolution \textit{local} views along with the standard $224\times 224$ ones. 
Use $2\times224 + 6 \times96$ scheme, where with two \textit{global} $224 \times 224$ augmented views, six \textit{local} $96 \times 96$ views are sampled, giving eight views per image. 
In this case, we follow the \textit{``local-to-global''} correspondence idea~\cite{caron2020unsupervised,caron2021emerging} and use only global views as positives for both local and global anchor images. 
While the proposed method shows slightly lower results compared to clustering-based SwAV, it again improves in the case of $k$-NN classification (Table~\ref{tab:sota_multicrop}).

\begin{table}[]
    \centering
    \resizebox{\columnwidth}{!}{
		 \begin{tabular}[t]{lc|ccc|cc}
			    \toprule 
			      \multirow{2}{*}{Method} & \multirow{2}{*}{Views} & \multicolumn{3}{c}{$k$-NN Evaluation} & \multicolumn{2}{c}{Linear Eval.} \\
			      &  & $k$=1 &  $k$=10 &  $k$=20 & Top-1 & Top-5  \\
			     \midrule
			     SimCLR & 2$\times$224 & - & - & - & 64.3 & - \\
			     \midrule
			     SimCLR$\ddagger$ & 2$\times$224 & 46.0 & 51.5 & 51.9 & 65.7 & 86.7 \\
			     SimCLR$\ddagger$ & 3$\times$224 & 44.7 & 50.0 & 50.6 & 65.8 & 86.8  \\
                  SimCLR$\ddagger$ & 4$\times$224 & 46.3 & 52.1 & 52.6 & 66.5 & 87.1 \\
			     SimCLR$\ddagger$ & 2$\times$224+6$\times$96 & 46.6   & 51.4 & 52.0 & 67.2 & 87.7 \\
			     \midrule
			    \rowcolor{almond} GroCo (ours) & 2$\times$224 & 55.3 & 60.3 & 60.5 & 69.2 & 88.4 \\
			      \rowcolor{almond} GroCo (ours) & 3$\times$224 & 55.8 & 61.2 & 61.6 & 69.5 & 88.8 \\
                  \rowcolor{almond} GroCo (ours) & 4$\times$224 & \underline{56.4} & \underline{61.5} & \underline{61.8} & \underline{69.6} & \underline{88.9} \\
    \rowcolor{almond} GroCo (ours) &  2$\times$224+6$\times$96 &  \textbf{57.2} & \textbf{62.0} & \textbf{62.3} & \textbf{71.8} & \textbf{90.4}\\
			\bottomrule
		\end{tabular}
		}
  \vspace{-1mm}
  \caption{\textbf{Comparison to SimCLR as a contrastive baseline on ImageNet.} \textit{Backbone=Resnet50, \#epochs=100, batch size=1024.} $\ddagger$~denotes our reproduction. \label{tab:simclr}}
  \vspace{-2mm}
\end{table}

\begin{table}[]
    \centering
    \resizebox{\columnwidth}{!}{
		 \begin{tabular}[t]{lc|ccc|cc}
			    \toprule 
			      \multirow{2}{*}{Method} & \multirow{2}{*}{Views} & \multicolumn{3}{c}{$k$-NN Evaluation} & \multicolumn{2}{c}{Linear Eval.} \\
			      &  & $k$=1 &  $k$=10 &  $k$=20 & Top-1 & Top-5  \\
			     \midrule
			     MoCo v3 & 2$\times$224 & 61.5 & 66.6 & 66.8 & \textbf{70.9} & \textbf{90.2} \\
			    \rowcolor{almond} GroCo (ours) & 2$\times$224 & \textbf{63.1} & \textbf{67.6} & \textbf{67.5} & 70.8 & 89.7 \\
            \bottomrule
		\end{tabular}
		}
  \vspace{-1mm}
  \caption{\textbf{Comparison to MoCo v3 as a contrastive baseline on ImageNet.} \textit{Backbone=ViT-Small, \#epochs=100, batch size=1024.} \label{tab:moco}}
  \vspace{-2mm}
\end{table}

\begin{table}[]
    \centering
    \resizebox{1.0\linewidth}{!}{
		 \begin{tabular}[t]{l|ccc|cc}
			    \toprule 
			    & \multicolumn{3}{c}{$k$-NN Evaluation} & \multicolumn{2}{c}{Linear Probing} \\
			      & $k$=1 &  $k$=10 &  $k$=20 & Top-1 & Top-5  \\
			     \midrule
                  Triplet Loss (margin=0.8) & 46.8 & 52.5 & 52.8 & 63.9 & 85.4 \\
                  Triplet Loss (margin=1.6) & 47.9 & 53.4 & 53.7 & 64.2 & 85.3 \\
                 Triplet Loss (margin=$+\infty$) & 47.9 & 53.3 & 53.8 & 64.3 & 85.3 \\
			     \rowcolor{almond} GroCo (ours) & \textbf{55.3} & \textbf{60.3} & \textbf{60.5} & \textbf{69.2} & \textbf{88.4} \\
			\bottomrule
		\end{tabular}
		}
  \vspace{-1mm}
  \caption{\textbf{Comparison on ImageNet with Triplet Loss.} \textit{Backbone=Resnet50, \#epochs=100, batch size=1024.} \label{tab:triplet}}
    \vspace{-3mm}
\end{table}

\vspace{-.3em}
\subsection{Comparison to Contrastive Loss}
\vspace{-.3em}

To evaluate the properties of the proposed method in a direct comparison with the pairwise contrastive loss formulation, we compare it further to the classical contrastive learning methods SimCLR~\cite{chen2020simple} and MoCo v3~\cite{mocov3} that use the popular InfoNCE loss~\cite{oord2018representation} as well as to the triplet loss.

\begin{table}[t]

\begingroup
\setlength{\tabcolsep}{2pt}
\begin{subtable}[t]{0.98\linewidth}
  \centering
    \resizebox{\columnwidth}{!}{
		 \begin{tabular}[t]{c|cc|cc|cc|cc}
			    \toprule 
			     Inverse temp. & \multicolumn{8}{c}{Number of negatives $N$}  \\
			     $\beta$  & \multicolumn{2}{c}{1}  & \multicolumn{2}{c}{5} & \multicolumn{2}{c}{10} & \multicolumn{2}{c}{20} \\
        \midrule
                 & $k$-NN & Lin.p. & $k$-NN & Lin.p. & $k$-NN & Lin.p. & $k$-NN & Lin.p. \\
                  
			     0.062 & 52.1 & 64.5 & -    &  -   & -    &  -   \\
			     0.125 & 52.2 & 64.5 & 59.2   & 68.7 & -    &  -   \\
			     0.25  & 52.1 & 64.4 & 59.6   & 68.6 & -    &  -   & -    &  -   \\
			     0.5   &    - &    - & 59.4   & 68.4 & 60.1 & 69.1 & -    &  -   \\
			     1     &    - &    - & 59.2   & 68.1 & \cellcolor{almond}\textbf{60.5} & \cellcolor{almond}\textbf{69.2} & 54.6 & 65.5 \\
			     2     &    - &    - & - & -    & 60.2 & 68.7 & 55.9 & 66.0 \\
			     4     &    - &    - & - & -    & -    & -    & 58.6 & 67.5 \\
			     8     &    - &    - & - & -    & -    & -    & 59.0 & 68.0 \\
			     16    &    - &    - & - & -    & -    & -    & 53.0 & 65.8 \\
			\bottomrule
		\end{tabular}
		}
  \caption{\textbf{An inverse temperature $\beta$, a number of negatives $N$.} \label{tab:temp_neg}}
\end{subtable} \\
\vspace{-2mm}
\endgroup

\begin{subtable}[t]{0.98\linewidth}
		\centering
    \resizebox{0.9\linewidth}{!}{
		 \begin{tabular}[t]{c|ccc|cc}
			    \toprule 
			      & \multicolumn{3}{c}{$k$-NN Evaluation} & \multicolumn{2}{c}{Linear Probing} \\
			      & $k$=1 &  $k$=10 &  $k$=20 & Top-1 & Top-5  \\
			     \midrule
			     Randomly ordered &  54.4 & 59.2 & 59.5 & 68.9 & 88.2 \\
        \rowcolor{almond} Pre-ordered & \textbf{55.3} & \textbf{60.3} & \textbf{60.5} & \textbf{69.2} & \textbf{88.4} \\
			\bottomrule
		\end{tabular}
		}
  \caption{\textbf{Pre-ordering in the groups.} \label{tab:preorder}}
	\end{subtable} \\
 \vspace{-2mm}

\begin{subtable}[t]{0.98\linewidth}
		\centering
    \resizebox{0.8\linewidth}{!}{
		 \begin{tabular}[t]{l|ccc|cc}
			    \toprule 
			      \multirow{2}{*}{Batch Size} & \multicolumn{3}{c}{$k$-NN Evaluation} & \multicolumn{2}{c}{Linear Probing} \\
			      & $k$=1 &  $k$=10 & $k$=20 & Top-1 & Top-5  \\
			     \midrule
			     256 & 52.4 & 57.2 & 57.2 & 67.8 & 88.1 \\
			     512 &  53.1 & 57.9 & 58.2 & 68.2 & 88.0 \\
			     \rowcolor{almond}1024 & \textbf{55.3} & \textbf{60.3} & \textbf{60.5} & \textbf{69.2} & \textbf{88.4}  \\
			\bottomrule
		\end{tabular}
		}
  \caption{\textbf{Batch size.} \label{tab:batch_size}}
  \vspace{-2mm}
\end{subtable} \\

\begin{subtable}[t]{0.98\linewidth}
		\centering
    \resizebox{1\linewidth}{!}{
		 \begin{tabular}[t]{l|ccc|cc}
			    \toprule 
			      \multirow{2}{*}{Batch Sampling} & \multicolumn{3}{c}{$k$-NN Evaluation} & \multicolumn{2}{c}{Linear Probing} \\
			      & $k$=1 &  $k$=10 & $k$=20 & Top-1 & Top-5  \\
			     \midrule
                  \rowcolor{almond} Regular & \textbf{55.3} & \textbf{60.3} & \textbf{60.5} & \textbf{69.2} & \textbf{88.4}  \\
                  With many false negatives &  51.5 & 56.6 & 56.9 & 67.6 & 87.7 \\
			\bottomrule
		\end{tabular}
		}
  \caption{\textbf{Sensitivity to false negatives in a batch.} 
  \vspace{-1mm}
  \label{tab:false_negatives}}
\end{subtable} 

\caption{\textbf{Ablation Experiments.} For (a), we report $k$-NN performance with $k$=20, and linear probing Top-1, denoted as ``Lin.p''.  The best results are bolded. Options used to obtain the main results are highlighted.
\textit{Backbone=Resnet50, Views=2$\times$224, \#epochs=100.} \label{tab:ablation}}
  \vspace{-3mm}
\end{table}

\myparagraph{Contrastive Loss.} 
First, we compare the performance to the SimCLR method in Table~\ref{tab:simclr}, where we also analyze the properties of losses to align local neighborhood based on more positive data points. To ensure an identical setting for both methods, SimCLR is reproduced with a 3-layer MLP projection head~\cite{chen2020big}.
Since SimCLR originally uses only two augmentations per image, we extend it to a group of positives by applying contrastive loss between all possible positive pairs (see supplementary material). 
While outperforming the SimCLR baseline by $+3\%$ (Top-$1$) in linear probing, the proposed method also advances in $k$-NN evaluation by more than $+8\%$ ($k=20$), demonstrating that the loss helps to learn better representation not only in terms of linear separability but also in terms of local structure.   %
Considering the multi-crop scenario, SimCLR shows mixed results from utilizing more views. While it benefits in linear evaluation, performance slightly decreases in $k$-NN for $k\in\{10,20\}$. The proposed method profits from utilizing more positives in both evaluations, resulting in~$+1.7\%$ in $k$-NN ($k=1$), and $+2.4\%$ in linear evaluation (Top-$1$). %
We further compare performance with the more recent and better-engineered MoCo~v3 approach~\cite{mocov3} with a ViT-Small~\cite{dosovitskiy2020image} backbone in Table~\ref{tab:moco}. MoCo v3 additionally utilizes a predictor and a momentum encoder that incur an additional computational cost but are beneficial for contrastive training~\cite{mocov3,caron2021emerging}. We plugin our proposed GroCo loss in MoCo~v3 setup (with the same architecture, optimizer, etc.) and perform minimal hyperparameter tuning. Table~\ref{tab:moco} shows that without any tips and tricks the proposed method outperforms MoCo~v3 in all $k$-NN metrics and has a comparable performance in Top-1 linear probing.

\begin{figure*}
\centering
\hfill
\begin{subfigure}{0.465\linewidth}
\centering
    \includegraphics[width=1\linewidth]{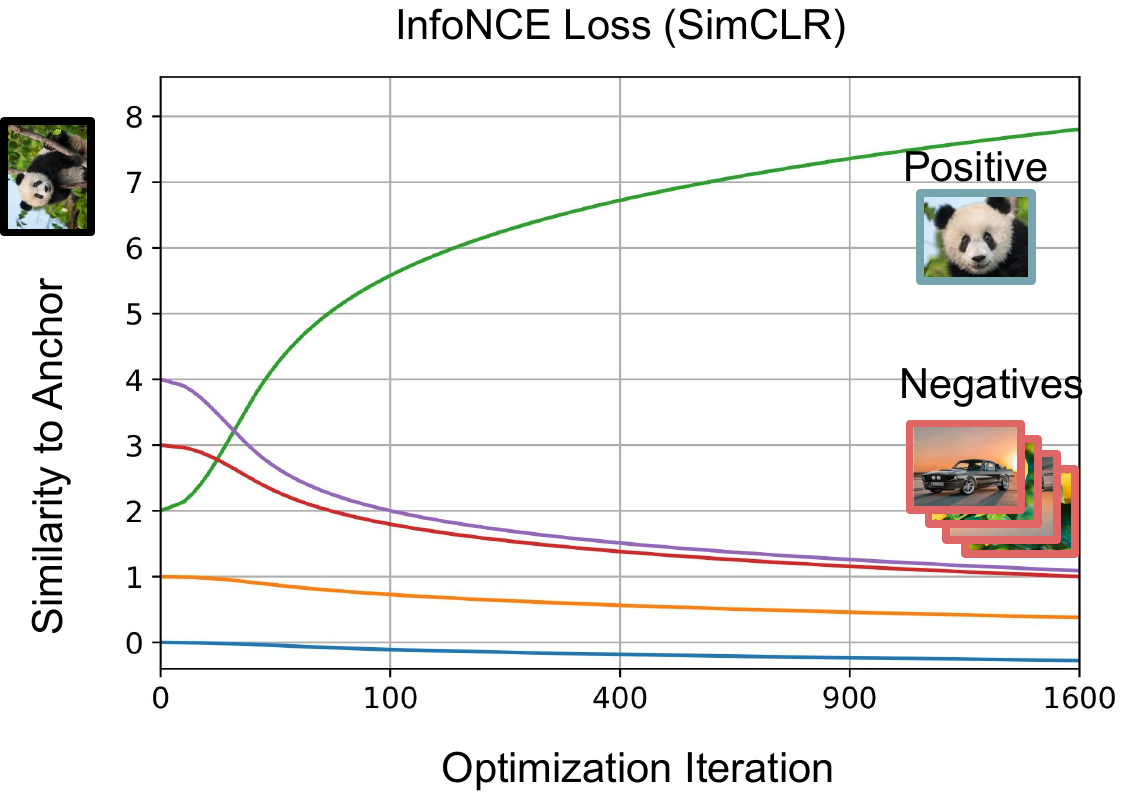}
\end{subfigure}%
\hfill\hfill
\begin{subfigure}{0.465\linewidth}
\centering
    \includegraphics[width=1\linewidth]{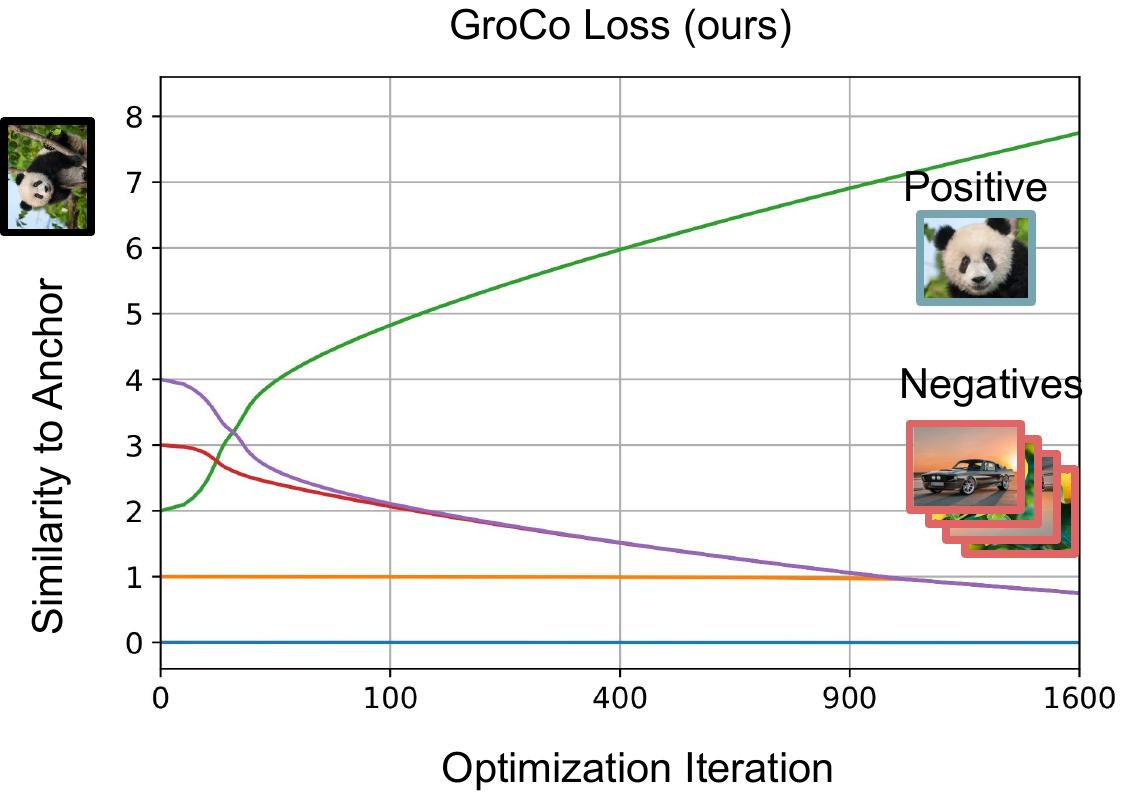}
\end{subfigure}%
\hfill
\caption{A toy experiment where we optimize five real variables, treating one of them as positive similarity and the other four as negatives, with both the GroCo loss and contrastive InfoNCE loss for multiple iterations.  
While the InfoNCE loss minimizes all negatives, the GroCo loss works behaves more similar to a margin optimization: negatives that further away from the border are optimized to a smaller degree. 
\label{fig:loss_optimization}}
\vspace{-2.5mm}
\end{figure*}

\myparagraph{Triplet Loss.} 
We also compare the group ordering loss with the triplet loss formulation $L=\max{(d_i^p - d_j^n + r, 0)}$, where $r$ is a margin (Table~\ref{tab:triplet}). 
For a fair comparison, we consider all positive and the 10 strongest negative samples and evaluate different margin parameters. Here, sorting is superior to a triplet loss with hard margin selection. %

\subsection{Ablation Study}
\label{sec:ablation}

\myparagraph{Inverse Temperature, Number of Negatives.}
Table~\ref{tab:temp_neg} shows the influence of the number of nearest neighbor negatives $N$ used in the loss, as well as the value of the inverse temperature parameter $\beta$ (Equation~\ref{eq:softswap}). 
We observe that usage of too many negatives is not be beneficial for the model. 
Since our loss focuses on negatives that are sorted incorrectly, increasing the number of negatives at some point does not bring any new learning signal (because more distant samples are unlikely to be sorted incorrectly). 
However, a larger $N$ results in more steps of the sorting network, increasing the degree of relaxation. 
Using a larger inverse temperature $\beta$ (leading to a lower degree of relaxation in the swap operation) we can gain some performance; however, the variance of the gradients is larger with a larger $\beta$, which is not beneficial for optimization. We found  $N=10$ and $\beta = 1$ to be an efficient configuration for this setting.

\myparagraph{Pre-ordering.} 
In Table~\ref{tab:preorder}, we analyze the impact of pre-ordering elements within negative and positive groups before forwarding them to the sorting network. We find that it achieves good performance even without pre-ordering, but also that pre-ordering further strengthens the method.

\myparagraph{Batch Size.} 
A large batch size can be an important factor in obtaining good performance for many self-supervised learning methods. 
{Results in Table~\ref{tab:batch_size} show that our method also benefits from a large batch size, which we attribute to utilizing stronger negatives from a larger batch.}

\myparagraph{False negatives.} 
{To assess the sensitivity of proposed methods to the false negatives, we artificially sample a batch in a way that for
each instance, there are three more instances of the same class on average (acting as false negatives). In Table~\ref{tab:false_negatives}, we observe a performance decline in this scenario ("With many false negatives"). Given GroCo's implicit emphasis on strong negatives, enhancing its robustness to false negatives would require further adjustments, which we leave for future work. }

\myparagraph{Training Time.} We also consider the training time of our model compared to SimCLR baseline. To eliminate the influence of distributed training, we measure the average time of training iteration on one GPU. We find that the iteration time of both models is comparable, 514ms for SimCLR vs 526ms for the proposed methods for a batch size of 128.

{
\myparagraph{Optimization of Negatives.} 
To better understand  the rationale behind GroCo's superiority in $k$-NN tasks, we analyze the difference between the GroCo loss and contrastive InfoNCE loss (used in SimCLR) in structuring the embedding space. 
We conduct a toy experiment where we optimize five real variables treating one of them as positive similarity and the other four as negative similarities with both losses for multiple iterations. We demonstrate the optimization process in~Figure~\ref{fig:loss_optimization}. 
Although both losses elevate positive similarities and lower negative similarities, they differ in the optimization of the negatives. 
While the InfoNCE loss minimizes all negatives (even pushing the blue curve substantially below zero), the GroCo loss works behaves more similar to a margin optimization: negatives that are not on the border to the positive are optimized to a smaller degree (the blue curve is pushed only slightly below zero). 
This highlights GroCo's focus on neighborhood optimization.
}

\section{Conclusion}
In this paper, an alternative approach to the common pairwise contrastive learning formulation is proposed. The group ordering constraints consider positives and negatives as groups and enforce the group of positives to be closer to the anchor image than the negative group. 
To enforce these constraints, recent progress in the context of differentiable sorting approaches are leveraged to formulate a group ordering loss based on the given sorting supervision. 
Our evaluation shows that the proposed framework, does not only compete with current contrastive loss baselines, but actually outperforms standard contrastive learning in many settings with regards to $k$-NN-based metrics. 
\subsection*{Acknowledgements}
\small
\noindent Nina Shvetsova is supported by German Federal Ministry of Education and Research (BMBF) project STCL - 01IS22067.

{\small
\bibliographystyle{ieee_fullname}
\bibliography{egbib}

\begin{thebibliography}{10}\itemsep=-1pt

\bibitem{akbari2021vatt}
Hassan Akbari, Liangzhe Yuan, Rui Qian, Wei-Hong Chuang, Shih-Fu Chang, Yin
  Cui, and Boqing Gong.
\newblock Vatt: Transformers for multimodal self-supervised learning from raw
  video, audio and text.
\newblock {\em NeurIPS}, 2021.

\bibitem{asano2019self}
YM Asano, C Rupprecht, and A Vedaldi.
\newblock Self-labelling via simultaneous clustering and representation
  learning.
\newblock In {\em ICLR}, 2020.

\bibitem{bachman2019learning}
Philip Bachman, R~Devon Hjelm, and William Buchwalter.
\newblock Learning representations by maximizing mutual information across
  views.
\newblock {\em NeurIPS}, 32, 2019.

\bibitem{bardes2021vicreg}
Adrien Bardes, Jean Ponce, and Yann LeCun.
\newblock Vicreg: Variance-invariance-covariance regularization for
  self-supervised learning.
\newblock In {\em ICLR}, 2022.

\bibitem{Blondel2020-FastSorting}
Mathieu Blondel, Olivier Teboul, Quentin Berthet, and Josip Djolonga.
\newblock {Fast Differentiable Sorting and Ranking}.
\newblock In {\em ICML}, 2020.

\bibitem{bossard2014food}
Lukas Bossard, Matthieu Guillaumin, and Luc~Van Gool.
\newblock Food-101--mining discriminative components with random forests.
\newblock In {\em ECCV}, 2014.

\bibitem{burges2006learning}
Christopher Burges, Robert Ragno, and Quoc Le.
\newblock Learning to rank with nonsmooth cost functions.
\newblock {\em NeurIPS}, 2006.

\bibitem{burges2005learning}
Chris Burges, Tal Shaked, Erin Renshaw, Ari Lazier, Matt Deeds, Nicole
  Hamilton, and Greg Hullender.
\newblock Learning to rank using gradient descent.
\newblock In {\em ICML}, 2005.

\bibitem{cao2007learning}
Zhe Cao, Tao Qin, Tie-Yan Liu, Ming-Feng Tsai, and Hang Li.
\newblock Learning to rank: from pairwise approach to listwise approach.
\newblock In {\em ICML}, 2007.

\bibitem{caron2018deep}
Mathilde Caron, Piotr Bojanowski, Armand Joulin, and Matthijs Douze.
\newblock Deep clustering for unsupervised learning of visual features.
\newblock In {\em ECCV}, 2018.

\bibitem{caron2020unsupervised}
Mathilde Caron, Ishan Misra, Julien Mairal, Priya Goyal, Piotr Bojanowski, and
  Armand Joulin.
\newblock Unsupervised learning of visual features by contrasting cluster
  assignments.
\newblock {\em NeurIPS}, 2020.

\bibitem{caron2021emerging}
Mathilde Caron, Hugo Touvron, Ishan Misra, Herv{\'e} J{\'e}gou, Julien Mairal,
  Piotr Bojanowski, and Armand Joulin.
\newblock Emerging properties in self-supervised vision transformers.
\newblock In {\em ICCV}, 2021.

\bibitem{Carr21Audio}
Andrew~N. Carr, Quentin Berthet, Mathieu Blondel, Olivier Teboul, and Neil
  Zeghidour.
\newblock Self-supervised learning of audio representations from permutations
  with differentiable ranking.
\newblock {\em IEEE Signal Processing Letters}, 28, 2021.

\bibitem{chen2020simple}
Ting Chen, Simon Kornblith, Mohammad Norouzi, and Geoffrey Hinton.
\newblock A simple framework for contrastive learning of visual
  representations.
\newblock In {\em ICML}, 2020.

\bibitem{chen2020big}
Ting Chen, Simon Kornblith, Kevin Swersky, Mohammad Norouzi, and Geoffrey~E
  Hinton.
\newblock Big self-supervised models are strong semi-supervised learners.
\newblock {\em NeurIPS}, 33, 2020.

\bibitem{chen2020improved}
Xinlei Chen, Haoqi Fan, Ross Girshick, and Kaiming He.
\newblock Improved baselines with momentum contrastive learning.
\newblock {\em arXiv preprint arXiv:2003.04297}, 2020.

\bibitem{chen2021exploring}
Xinlei Chen and Kaiming He.
\newblock Exploring simple siamese representation learning.
\newblock In {\em CVPR}, 2021.

\bibitem{mocov3}
X. Chen, S. Xie, and K. He.
\newblock An empirical study of training self-supervised vision transformers.
\newblock In {\em ICCV}, Los Alamitos, CA, USA, oct 2021. IEEE Computer
  Society.

\bibitem{cimpoi2014describing}
Mircea Cimpoi, Subhransu Maji, Iasonas Kokkinos, Sammy Mohamed, and Andrea
  Vedaldi.
\newblock Describing textures in the wild.
\newblock In {\em CVPR}, 2014.

\bibitem{cordonnier2021differentiable}
Jean-Baptiste Cordonnier, Aravindh Mahendran, Alexey Dosovitskiy, Dirk
  Weissenborn, Jakob Uszkoreit, and Thomas Unterthiner.
\newblock Differentiable patch selection for image recognition.
\newblock In {\em CVPR}, 2021.

\bibitem{cuturi2019differentiable}
Marco Cuturi, Olivier Teboul, and Jean-Philippe Vert.
\newblock Differentiable ranking and sorting using optimal transport.
\newblock {\em NeurIPS}, 2019.

\bibitem{dosovitskiy2020image}
Alexey Dosovitskiy, Lucas Beyer, Alexander Kolesnikov, Dirk Weissenborn,
  Xiaohua Zhai, Thomas Unterthiner, Mostafa Dehghani, Matthias Minderer, Georg
  Heigold, Sylvain Gelly, Jakob Uszkoreit, and Neil Houlsby.
\newblock An image is worth 16x16 words: Transformers for image recognition at
  scale.
\newblock In {\em ICLR}, 2021.

\bibitem{dwibedi2021little}
Debidatta Dwibedi, Yusuf Aytar, Jonathan Tompson, Pierre Sermanet, and Andrew
  Zisserman.
\newblock With a little help from my friends: Nearest-neighbor contrastive
  learning of visual representations.
\newblock In {\em CVPR}, 2021.

\bibitem{ermolov2021whitening}
Aleksandr Ermolov, Aliaksandr Siarohin, Enver Sangineto, and Nicu Sebe.
\newblock Whitening for self-supervised representation learning.
\newblock In {\em ICML}, 2021.

\bibitem{everingham2010pascal}
Mark Everingham, Luc Van~Gool, Christopher~KI Williams, John Winn, and Andrew
  Zisserman.
\newblock The pascal visual object classes (voc) challenge.
\newblock {\em IJCV}, 88(2):303--338, 2010.

\bibitem{fei2004learning}
Li Fei-Fei, Rob Fergus, and Pietro Perona.
\newblock Learning generative visual models from few training examples: An
  incremental bayesian approach tested on 101 object categories.
\newblock In {\em CVPR}, 2004.

\bibitem{grill2020bootstrap}
Jean-Bastien Grill, Florian Strub, Florent Altch{\'e}, Corentin Tallec, Pierre
  Richemond, Elena Buchatskaya, Carl Doersch, Bernardo Avila~Pires, Zhaohan
  Guo, Mohammad Gheshlaghi~Azar, et~al.
\newblock Bootstrap your own latent-a new approach to self-supervised learning.
\newblock {\em NeurIPS}, 33, 2020.

\bibitem{grover2019stochastic}
Aditya Grover, Eric Wang, Aaron Zweig, and Stefano Ermon.
\newblock {Stochastic Optimization of Sorting Networks via Continuous
  Relaxations}.
\newblock In {\em ICLR}, 2019.

\bibitem{hazimeh2021dselect}
Hussein Hazimeh, Zhe Zhao, Aakanksha Chowdhery, Maheswaran Sathiamoorthy, Yihua
  Chen, Rahul Mazumder, Lichan Hong, and Ed Chi.
\newblock Dselect-k: Differentiable selection in the mixture of experts with
  applications to multi-task learning.
\newblock {\em NeurIPS}, 34, 2021.

\bibitem{he2020moco}
Kaiming He, Haoqi Fan, Yuxin Wu, Saining Xie, and Ross Girshick.
\newblock Momentum contrast for unsupervised visual representation learning.
\newblock {\em CVPR}, 2012.

\bibitem{he2016deep}
Kaiming He, Xiangyu Zhang, Shaoqing Ren, and Jian Sun.
\newblock Deep residual learning for image recognition.
\newblock In {\em CVPR}, 2016.

\bibitem{huynh2022boosting}
Tri Huynh, Simon Kornblith, Matthew~R Walter, Michael Maire, and Maryam
  Khademi.
\newblock Boosting contrastive self-supervised learning with false negative
  cancellation.
\newblock In {\em WACV}, 2022.

\bibitem{ioffe2015batch}
Sergey Ioffe and Christian Szegedy.
\newblock Batch normalization: Accelerating deep network training by reducing
  internal covariate shift.
\newblock In {\em ICML}, 2015.

\bibitem{knuth1998sorting}
Donald~E. Knuth.
\newblock {\em The Art of Computer Programming, Volume 3: Sorting and Searching
  (2nd Ed.)}.
\newblock Addison Wesley, 1998.

\bibitem{krause2013collecting}
Jonathan Krause, Jia Deng, Michael Stark, and Li Fei-Fei.
\newblock Collecting a large-scale dataset of fine-grained cars.
\newblock 2013.

\bibitem{krizhevsky2009learning}
Alex Krizhevsky, Geoffrey Hinton, et~al.
\newblock Learning multiple layers of features from tiny images.
\newblock 2009.

\bibitem{lee2021differentiable}
Hyunsung Lee, Sangwoo Cho, Yeongjae Jang, Jaekwang Kim, and Honguk Woo.
\newblock Differentiable ranking metric using relaxed sorting for top-k
  recommendation.
\newblock {\em IEEE Access}, 9, 2021.

\bibitem{loshchilov2016sgdr}
Ilya Loshchilov and Frank Hutter.
\newblock {SGDR:} stochastic gradient descent with warm restarts.
\newblock In {\em ICLR}, 2017.

\bibitem{maji2013fine}
Subhransu Maji, Esa Rahtu, Juho Kannala, Matthew Blaschko, and Andrea Vedaldi.
\newblock Fine-grained visual classification of aircraft.
\newblock {\em arXiv preprint arXiv:1306.5151}, 2013.

\bibitem{miech20endtoend}
Antoine Miech, Jean-Baptiste Alayrac, Lucas Smaira, Ivan Laptev, Josef Sivic,
  and Andrew Zisserman.
\newblock {E}nd-to-{E}nd {L}earning of {V}isual {R}epresentations from
  {U}ncurated {I}nstructional {V}ideos.
\newblock In {\em CVPR}, 2020.

\bibitem{nilsback2008automated}
Maria-Elena Nilsback and Andrew Zisserman.
\newblock Automated flower classification over a large number of classes.
\newblock In {\em 2008 Sixth Indian Conference on Computer Vision, Graphics \&
  Image Processing}, pages 722--729. IEEE, 2008.

\bibitem{oord2018representation}
Aaron van~den Oord, Yazhe Li, and Oriol Vinyals.
\newblock Representation learning with contrastive predictive coding.
\newblock {\em arXiv preprint arXiv:1807.03748}, 2018.

\bibitem{parkhi2012cats}
Omkar~M Parkhi, Andrea Vedaldi, Andrew Zisserman, and CV Jawahar.
\newblock Cats and dogs.
\newblock In {\em CVPR}. IEEE, 2012.

\bibitem{petersen2022thesis}
Felix Petersen.
\newblock {\em Learning with Differentiable Algorithms}.
\newblock PhD thesis, Universit\"at Konstanz, 2022.

\bibitem{petersen2021diffsort}
Felix Petersen, Christian Borgelt, Hilde Kuehne, and Oliver Deussen.
\newblock {Differentiable Sorting Networks for Scalable Sorting and Ranking
  Supervision}.
\newblock In {\em ICML}, 2021.

\bibitem{petersen2021learning}
Felix Petersen, Christian Borgelt, Hilde Kuehne, and Oliver Deussen.
\newblock Learning with algorithmic supervision via continuous relaxations.
\newblock {\em NeurIPS}, 2021.

\bibitem{petersen2022monotonic}
Felix Petersen, Christian Borgelt, Hilde Kuehne, and Oliver Deussen.
\newblock Monotonic differentiable sorting networks.
\newblock {\em ICLR}, 2022.

\bibitem{petersen2022differentiable}
Felix Petersen, Hilde Kuehne, Christian Borgelt, and Oliver Deussen.
\newblock Differentiable top-k classification learning.
\newblock In {\em ICML}, 2022.

\bibitem{prillo2020softsort}
Sebastian Prillo and Julian Eisenschlos.
\newblock Softsort: A continuous relaxation for the argsort operator.
\newblock In {\em ICML}, 2020.

\bibitem{radenovic2018revisiting}
Filip Radenovi{\'c}, Ahmet Iscen, Giorgos Tolias, Yannis Avrithis, and
  Ond{\v{r}}ej Chum.
\newblock Revisiting oxford and paris: Large-scale image retrieval
  benchmarking.
\newblock In {\em CVPR}, 2018.

\bibitem{radford2021learning}
Alec Radford, Jong~Wook Kim, Chris Hallacy, Aditya Ramesh, Gabriel Goh,
  Sandhini Agarwal, Girish Sastry, Amanda Askell, Pamela Mishkin, Jack Clark,
  et~al.
\newblock Learning transferable visual models from natural language
  supervision.
\newblock In {\em ICML}, 2021.

\bibitem{Radford2021CLIP}
Alec Radford, Jong~Wook Kim, Chris Hallacy, Aditya Ramesh, Gabriel Goh,
  Sandhini Agarwal, Girish Sastry, Amanda Askell, Pamela Mishkin, Jack Clark,
  Gretchen Krueger, and Ilya Sutskever.
\newblock Learning transferable visual models from natural language
  supervision.
\newblock In {\em ICML}, 2021.

\bibitem{robinson2020contrastive}
Joshua~David Robinson, Ching-Yao Chuang, Suvrit Sra, and Stefanie Jegelka.
\newblock Contrastive learning with hard negative samples.
\newblock In {\em ICLR}, 2021.

\bibitem{russakovsky2015imagenet}
Olga Russakovsky, Jia Deng, Hao Su, Jonathan Krause, Sanjeev Satheesh, Sean Ma,
  Zhiheng Huang, Andrej Karpathy, Aditya Khosla, Michael Bernstein, et~al.
\newblock Imagenet large scale visual recognition challenge.
\newblock {\em IJCV}, 115(3), 2015.

\bibitem{schroff2015facenet}
Florian Schroff, Dmitry Kalenichenko, and James Philbin.
\newblock Facenet: A unified embedding for face recognition and clustering.
\newblock In {\em CVPR}, 2015.

\bibitem{schuhmann2022laionb}
Christoph Schuhmann, Romain Beaumont, Richard Vencu, Cade~W Gordon, Ross
  Wightman, Mehdi Cherti, Theo Coombes, Aarush Katta, Clayton Mullis, Mitchell
  Wortsman, Patrick Schramowski, Srivatsa~R Kundurthy, Katherine Crowson,
  Ludwig Schmidt, Robert Kaczmarczyk, and Jenia Jitsev.
\newblock {LAION}-5b: An open large-scale dataset for training next generation
  image-text models.
\newblock In {\em NeurIPS Datasets and Benchmarks Track}, 2022.

\bibitem{shah2022max}
Anshul Shah, Suvrit Sra, Rama Chellappa, and Anoop Cherian.
\newblock Max-margin contrastive learning.
\newblock In {\em Proceedings of the AAAI Conference on Artificial
  Intelligence}, 2022.

\bibitem{swezey2021pirank}
Robin Swezey, Aditya Grover, Bruno Charron, and Stefano Ermon.
\newblock Pirank: Scalable learning to rank via differentiable sorting.
\newblock {\em NeurIPS}, 2021.

\bibitem{taylor2008softrank}
Michael Taylor, John Guiver, Stephen Robertson, and Tom Minka.
\newblock Softrank: optimizing non-smooth rank metrics.
\newblock In {\em Proceedings of the 2008 International Conference on Web
  Search and Data Mining}, pages 77--86, 2008.

\bibitem{tian2020makes}
Yonglong Tian, Chen Sun, Ben Poole, Dilip Krishnan, Cordelia Schmid, and
  Phillip Isola.
\newblock What makes for good views for contrastive learning?
\newblock {\em NeurIPS}, 2020.

\bibitem{xiao2010sun}
Jianxiong Xiao, James Hays, Krista~A Ehinger, Aude Oliva, and Antonio Torralba.
\newblock Sun database: Large-scale scene recognition from abbey to zoo.
\newblock In {\em CVPR}, 2010.

\bibitem{xiao2020should}
Tete Xiao, Xiaolong Wang, Alexei~A Efros, and Trevor Darrell.
\newblock What should not be contrastive in contrastive learning.
\newblock In {\em ICLR}, 2021.

\bibitem{xuan2020hard}
Hong Xuan, Abby Stylianou, Xiaotong Liu, and Robert Pless.
\newblock Hard negative examples are hard, but useful.
\newblock In {\em ECCV}, 2020.

\bibitem{ye2019unsupervised}
Mang Ye, Xu Zhang, Pong~C Yuen, and Shih-Fu Chang.
\newblock Unsupervised embedding learning via invariant and spreading instance
  feature.
\newblock In {\em CVPR}, 2019.

\bibitem{you2017large}
Yang You, Igor Gitman, and Boris Ginsburg.
\newblock Large batch training of convolutional networks.
\newblock {\em arXiv preprint arXiv:1708.03888}, 2017.

\bibitem{zbontar2021barlow}
Jure Zbontar, Li Jing, Ishan Misra, Yann LeCun, and St{\'e}phane Deny.
\newblock Barlow twins: Self-supervised learning via redundancy reduction.
\newblock In {\em ICML}, 2021.

\bibitem{zhan2021bi}
Fangneng Zhan, Yingchen Yu, Rongliang Wu, Kaiwen Cui, Aoran Xiao, Shijian Lu,
  and Ling Shao.
\newblock Bi-level feature alignment for versatile image translation and
  manipulation.
\newblock In {\em ECCV}, 2021.

\bibitem{zheng2021ressl}
Mingkai Zheng, Shan You, Fei Wang, Chen Qian, Changshui Zhang, Xiaogang Wang,
  and Chang Xu.
\newblock Ressl: Relational self-supervised learning with weak augmentation.
\newblock {\em NeurIPS}, 2021.

\end{thebibliography}
}
\clearpage

\addcontentsline{toc}{section}{Appendix} 
\appendix
\noindent{\Large\bf Supplementary Material}\\[1em]

\noindent In the supplementary material, we first discuss relations between the GroCo loss, the contrastive loss, and the triplet loss in Section~\ref{sec:disc_losses}. 
Then, we provide additional experimental evaluation results in Section~\ref{sec:add_exp_res} and a qualitative analysis in Section~\ref{sec:repres}. 
In Section~\ref{sec:odd_even}, we describe odd-even sorting networks.
Finally, we cover additional implementation details in Section~\ref{sec:supp_imp_detail}.

\section{Discussion of GroCo / Contrastive / Triplet Loss Relations}
\label{sec:disc_losses}

\begin{figure*}
\begin{subfigure}{0.33\linewidth}
\centering
    \caption*{$\log{(1 + \exp(- (d^n - d^p) / \tau))}$}
    \vspace{-3mm}
    \includegraphics[width=1\linewidth]{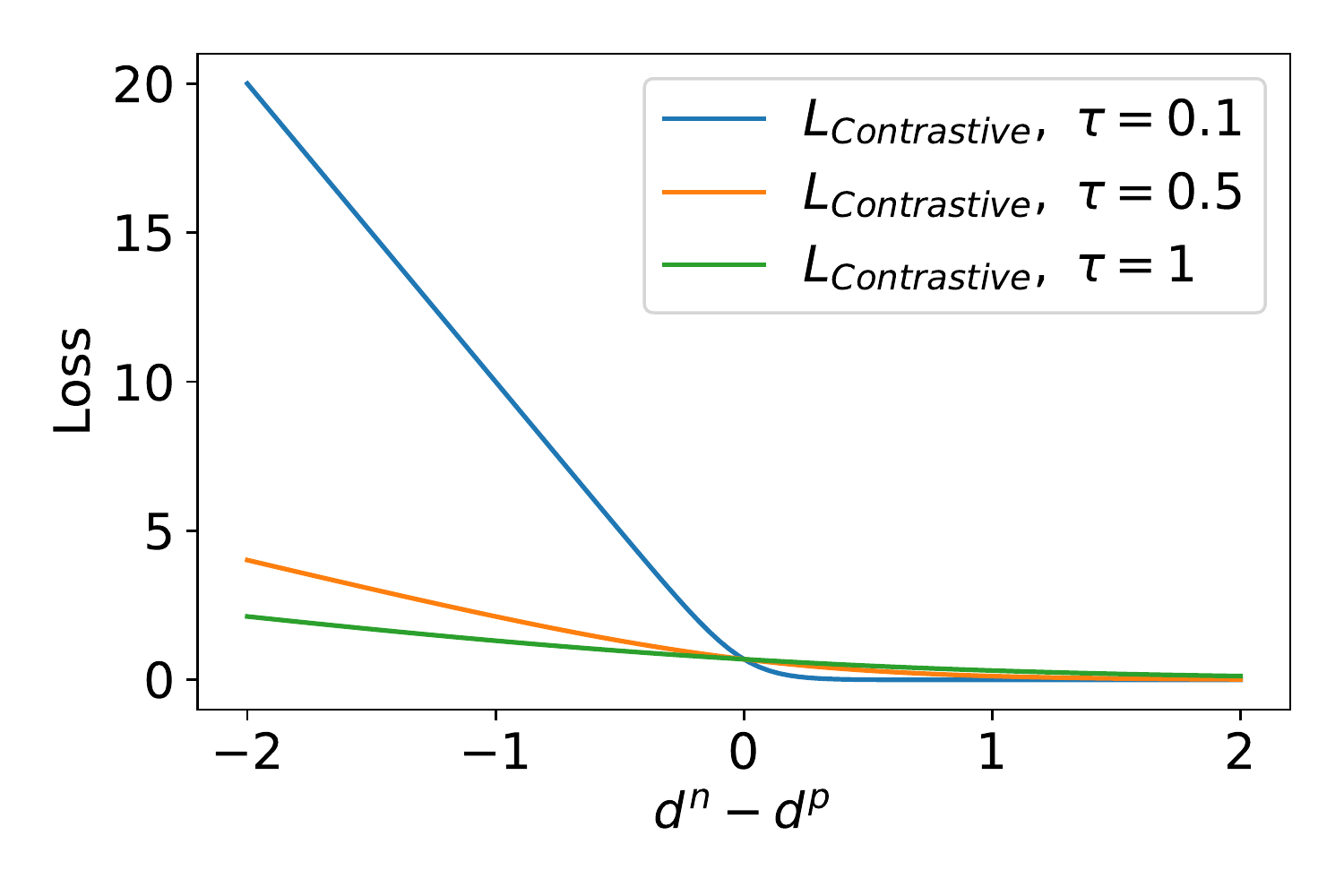}
    \caption{Contrastive InfoNCE loss}
    \label{fig:cont_loss}
\end{subfigure}%
\begin{subfigure}{0.33\linewidth}
\centering
    \caption*{$\max{(r - (d^n - d^p), 0)}$}
    \vspace{-3mm}
    \includegraphics[width=1\linewidth]{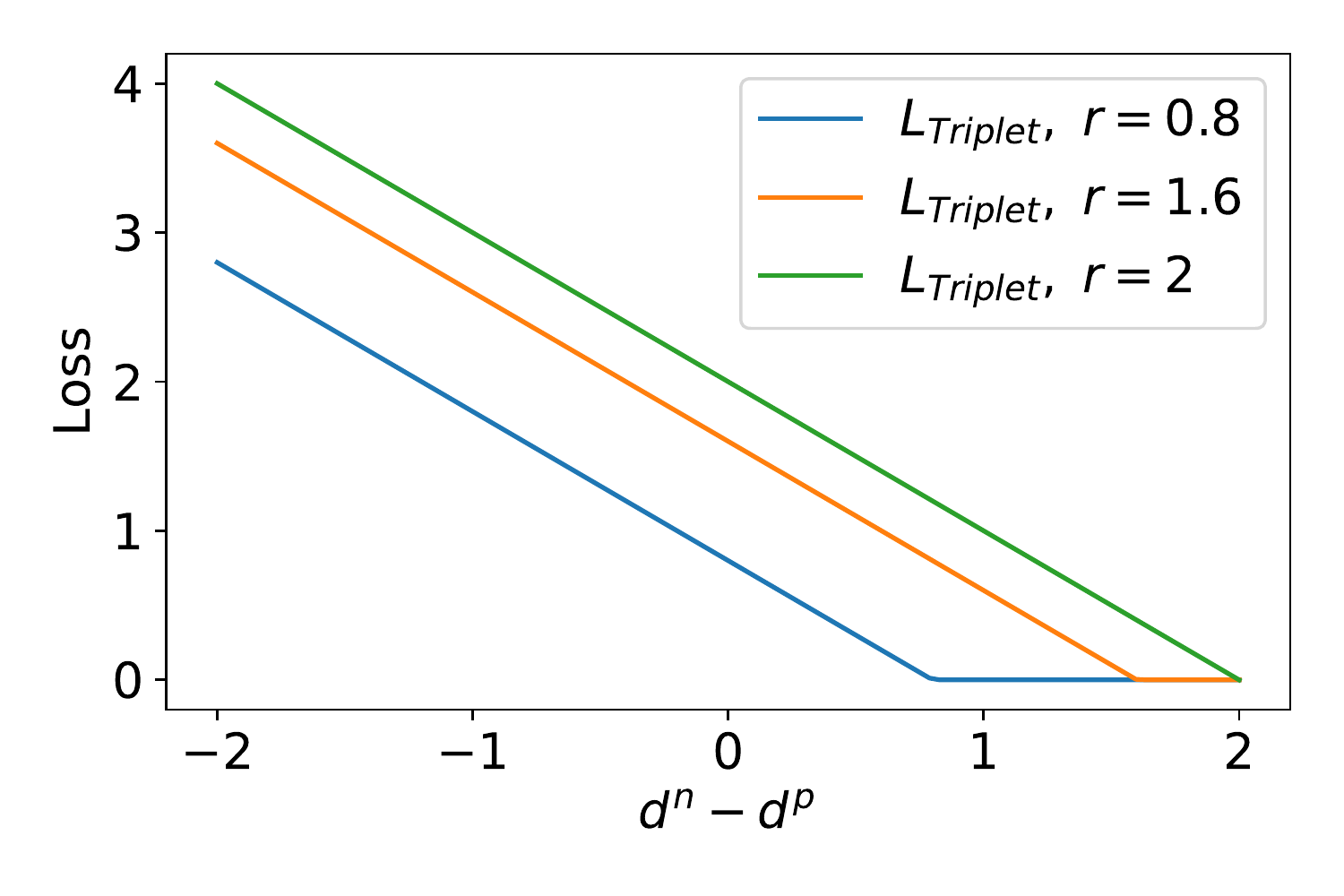}
    \caption{Triplet loss}
    \label{fig:trip_loss}
\end{subfigure}%
\begin{subfigure}{0.33\linewidth}
\centering
    \caption*{$- \log(\frac{1}{\pi} \arctan(\beta (d^n - d^p)) + 0.5)$
    }
    \vspace{-3mm}
    \includegraphics[width=1\linewidth]{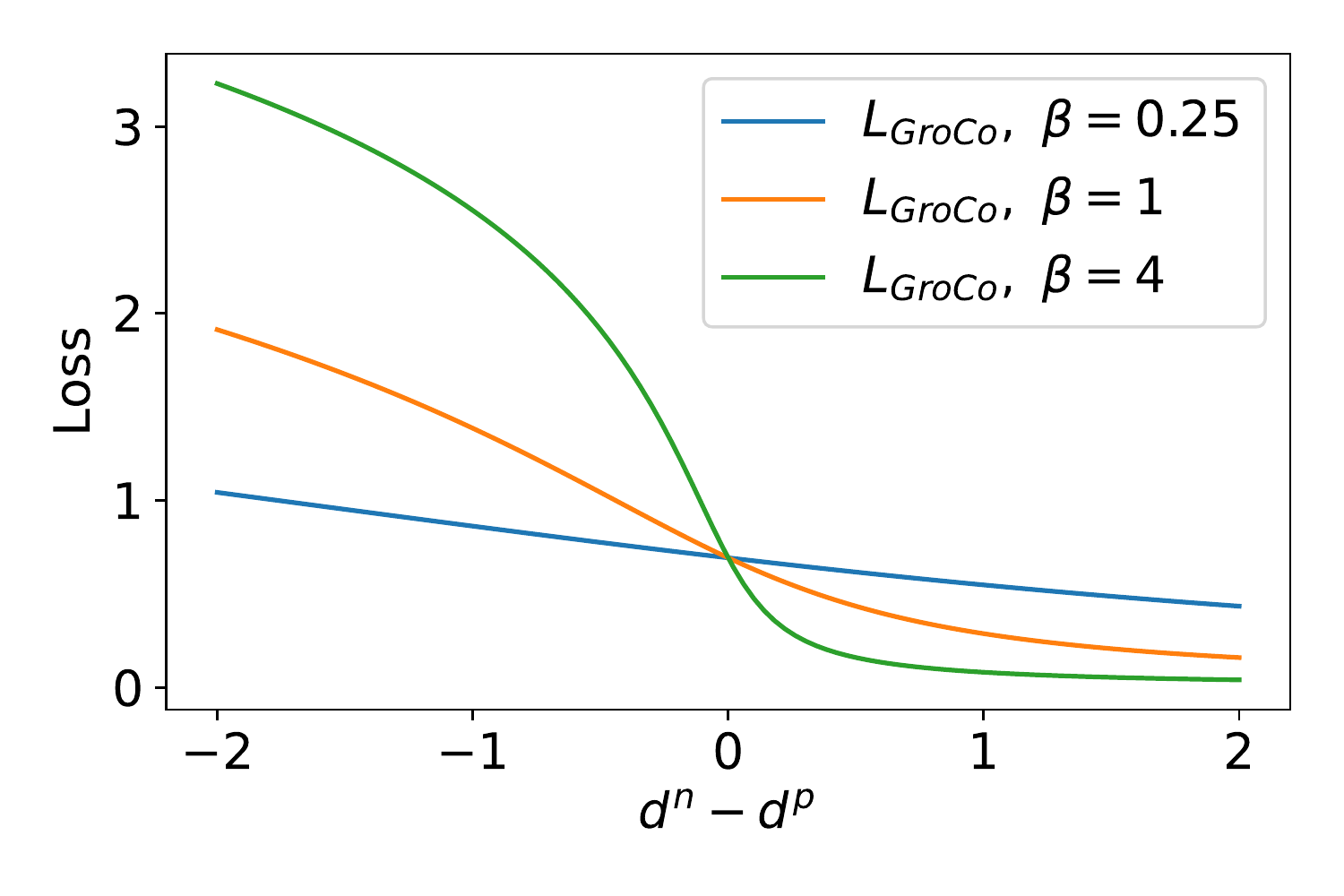}
    \caption{GroCo loss
    }
    \label{fig:groco_loss}
\end{subfigure}

\caption{Comparison of the contrastive loss, the triplet loss, and the GroCo loss in a simple scenario with only one positive example and one negative example for an anchor image.
We denote the distance from the anchor to the positive sample as $d^p$ and the distance from the anchor to the negative sample as $d^n$. 
We note that, in the simple case of only one positive and one negative, all three losses try to maximize the difference between the distances to the positive and negative examples $(d^n - d^p)$. The temperature $\tau$, the margin $r$, or the inverse temperature $\beta$ define the flatness of the loss curve depending on the difference $(d^n - d^p)$. 
\label{fig:loss_comp}}
\end{figure*}

In this section, we discuss the similarities and differences between the GroCo loss, the contrastive loss, and the triplet loss.
For comparison purposes, let's consider a simplified version of losses when there is only one positive example $x^p$ and one negative example $x^n$ for the anchor $x^a$. 
We denote the distance from the anchor $x^a$ to the positive sample  $x^p$ as 
$d^p = - \frac{{x^a}^\top x^p}{\left\lVert x^a\right\rVert \left\lVert x^p\right\rVert}$ 
and the distance from the anchor $x^a$ to the negative sample $x^n$  as
$d^n = - \frac{{x^a}^\top x^n}{\left\lVert x^a\right\rVert \left\lVert x^n\right\rVert}$. 
Then contrastive InfoNCE loss (with respect to the anchor $x^a$) is defined as:
\begin{equation}
\begin{aligned}
    L_{Contrastive} = - &\log{\frac{\exp(-d^p / \tau)}{\exp(-d^p / \tau) + \exp(-d^n / \tau)}} \\
            = &\log{(1 + \exp(- (d^n - d^p) / \tau))}
\end{aligned}
\end{equation}
where $\tau$ is a temperature hyperparameter~(Figure~\ref{fig:cont_loss}).

The triplet loss is defined as: 
\begin{equation}
\begin{aligned}
    L_{Triplet} & = \max{(d^p - d^n + r, 0)} = \\
      & = \max{(r - (d^n - d^p), 0)}
\end{aligned}
\end{equation}
where $r$ is a margin hyperparameter~(Figure~\ref{fig:trip_loss}).

For the GroCo loss, a permutation matrix $P \in R^{2\times2}$ corresponds to only one conditional swap operation and is defined as: 
\begin{equation}
\begin{split}
\begin{aligned}
    &P_{11} = P_{22} = f(d^n - d^p) = \frac{1}{\pi} \arctan(\beta (d^n - d^p)) + 0.5\,, \\
    &P_{12} = P_{21} = f(d^p - d^n) = \frac{1}{\pi} \arctan(\beta (d^p - d^n)) + 0.5 \\
\end{aligned}
\end{split}
\end{equation}
where $\beta$ is an inverse temperature. Therefore, the GroCo loss is defined as:
\begin{equation}
\begin{split}
\begin{aligned}
    L_{GroCo} &= \frac{1}{4}\Biggl(- 2 \log\Bigl(\frac{1}{\pi} \arctan(\beta (d^n - d^p)) + 0.5\Bigl) - \\
        - &2 \log\Bigl(1 - \frac{1}{\pi} \arctan(\beta (d^p - d^n)) - 0.5\Bigl)\Biggl) = \\
        &= - \log\Bigl(\frac{1}{\pi} \arctan(\beta (d^n - d^p)) + 0.5\Bigl)
\end{aligned}
\end{split}
\end{equation}
where $\beta$ is an inverse temperature hyperparameter~(Figure~\ref{fig:groco_loss}).

In Figure~\ref{fig:loss_comp}, 
we show the loss curves with different values of respective hyperparameters. We note that in this simplified example with only one positive and only one negative, all three losses try to maximize the difference between the distances to the positive and negative examples $(d^n - d^p)$. The temperature $\tau$, the margin $r$, or the inverse temperature $\beta$ define the flatness of the loss curve depending on the difference $(d^n - d^p)$. 

However, in the case with more negative/positive examples for the anchor image, different losses 
integrate information from multiple negatives/positives in different ways. 
For the triplet loss, there are various strategies to sample one positive example and one negative example for the anchor image~\cite{schroff2015facenet,xuan2020hard}. The complete loss is defined as the sum (or average) of the losses for the
chosen triplets $\sum_{ij}\max{(r - (d^n_i - d^p_j), 0)}$. On the other hand, the contrastive loss aggregates multiple negatives by contrasting the positive example to all negative examples, resulting in sum under logarithm: $\log{(1 + \sum_{i}\exp(- (d^n_i - d^p) / \tau))}$. 
In contrast to an explicit sum over a predefined number of negatives, the GroCo loss aggregates multiple positives and negatives via the permutation matrix, conditionally swapping neighboring elements, and later applies the group ordering supervision, enforcing the distance between positive and negative groups.

\section{Additional Experimental Results}
\label{sec:add_exp_res}

\noindent In this section, we provide additional experimental results:

\begin{table}[t]

\begingroup
\setlength{\tabcolsep}{3pt}
  \begin{subtable}[t]{0.98\linewidth}
		\centering
    \resizebox{1\linewidth}{!}{
		 \begin{tabular}[t]{l|ccc|cc}
			    \toprule 
			      \multirow{2}{*}{} & \multicolumn{3}{c}{$k$-NN Evaluation} & \multicolumn{2}{c}{Linear Probing} \\
			      & $k$=1 &  $k$=10 & $k$=20 & Top-1 & Top-5  \\
			     \midrule
                  InfoNCE (= SimCLR method) & 46.0 & 51.5 & 51.9 & 65.7 & 86.7 \\
                  InfoNCE + s. grad. & 46.1 & 51.3 & 51.8 & 65.6 & 86.7 \\ 
                  InfoNCE + top 10 neg. & \multicolumn{5}{c}{unstable training}  \\
			     InfoNCE + top 10 neg. + s. grad. & 49.5 & 54.6 & 55.0 & 66.0 & 86.7 \\
                  GroCo + top 10 neg. & \multicolumn{5}{c}{unstable training} \\
			     \rowcolor{almond}GroCo + top 10 neg. + s. grad. & \textbf{55.3} & \textbf{60.3} & \textbf{60.5} & \textbf{69.2} & \textbf{88.4}  \\
			\bottomrule
		\end{tabular}
		}
  \caption{\textbf{Usage of top-10 negatives, stop gradient operation} \label{tab:stop_grad}}
\end{subtable} \\
\endgroup

\begin{subtable}[t]{0.98\linewidth}
\centering
\resizebox{1\linewidth}{!}{
\begin{tabular}[t]{lcc|ccc|cc}
			    \toprule 
			      \multirow{2}{*}{Method} & \multirow{2}{*}{Views} & \multirow{2}{*}{Epochs} & \multicolumn{3}{c}{$k$-NN Evaluation} & \multicolumn{2}{c}{Linear Eval.} \\
			      &  &  & $k$=1 &  $k$=10 &  $k$=20 & Top-1 & Top-5  \\
			     \midrule
                   \rowcolor{almond} GroCo  & 2$\times$224 & 800 & 59.9 & 65.0 & 65.3 & 71.2 & 89.9 \\
                \rowcolor{almond} GroCo  &  2$\times$224+6$\times$96 & 800 & 60.8 & 65.7 & 66.1 & 73.9 & 91.6 \\
			\bottomrule
		\end{tabular}
}
\caption{\textbf{Longer training} \label{tab:longer_training}}
\end{subtable} \\

\begin{subtable}[t]{0.98\linewidth}
    \centering
        \resizebox{\columnwidth}{!}{
		 \begin{tabular}[t]{llc|ccc|cc}
			    \toprule 
			     \multirow{2}{*}{Method} & \multirow{2}{*}{Augmentations} & \multirow{2}{*}{Epochs} & \multicolumn{3}{c}{$k$-NN Evaluation} & \multicolumn{2}{c}{Linear Evaluation} \\
			      & & & $k$=1 &  $k$=10 &  $k$=20 & Top-1 & Top-5  \\
\midrule
\rowcolor{almond} SimCLR & as in SimCLR~\cite{caron2021emerging} & 100 & \textbf{46.0} & \textbf{51.5} & \textbf{51.9} & \textbf{65.7} & \textbf{86.7} \\
SimCLR & as in DINO~\cite{caron2021emerging} & 100 & 43.3 & 48.6 & 49.1 & 63.7 & 85.4 \\
\midrule
GroCo & as in SimCLR~\cite{caron2021emerging} & 100 & 54.0 & 59.0 & 59.4 & 68.4 & 88.3 \\
\rowcolor{almond} GroCo & as in DINO~\cite{caron2021emerging} & 100 & \textbf{55.3} & \textbf{60.3} & \textbf{60.5} & \textbf{69.2} & \textbf{88.4} \\
\midrule
GroCo & as in SimCLR~\cite{caron2021emerging} & 200 & 56.7 & 61.6 & 61.8 & 69.8 & 89.1\\
\rowcolor{almond} GroCo & as in DINO~\cite{caron2021emerging} & 200 & \textbf{57.7} & \textbf{62.4} & \textbf{62.7} & \textbf{70.4} & \textbf{89.5} \\
\midrule
GroCo & as in SimCLR~\cite{caron2021emerging} & 400 & 58.3 & 63.3 & 63.8 & 71.1 & 89.7 \\ 
\rowcolor{almond} GroCo & as in DINO~\cite{caron2021emerging} & 400 & 58.7 & 63.4 & 63.6 & 71.0 &89.7 \\
			\bottomrule
		\end{tabular}
		}
  \caption{\textbf{Augmentation strategy} \label{tab:augmentation}}
\end{subtable} \\

\begin{subtable}[t]{0.98\linewidth}
    \centering
         \resizebox{1\columnwidth}{!}{
		 \begin{tabular}[t]{cc|ccc|cc}
			    \toprule 
			      \multirow{2}{*}{Projection dim} & \multirow{2}{*}{Embedding dim} & \multicolumn{3}{c}{$k$-NN Evaluation} & \multicolumn{2}{c}{Linear Probing} \\
			      & &  $k$=1 &  $k$=10 &  $k$=20 & Top-1 & Top-5  \\
			     \midrule
			    128 & 2048 & 53.7 & 58.5 & 58.7 & 68.1 & 88.0  \\
			      512 & 2048 & 55.2 & 59.8 & 60.1 & 69.0 & \textbf{88.5}  \\
                    \rowcolor{almond} 2048 & 2048 & \textbf{55.3} & \textbf{60.3} & \textbf{60.5} & \textbf{69.2} & 88.4 \\
			\bottomrule
		\end{tabular}
        }
  \caption{\textbf{Projection dimentionality} \label{tab:proj_dim}}
\end{subtable} \\

\begin{subtable}[t]{0.98\linewidth}
		\centering
    \resizebox{0.9\columnwidth}{!}{
		 \begin{tabular}[t]{l|ccc|cc}
			    \toprule 
			      & \multicolumn{3}{c}{$k$-NN Evaluation} & \multicolumn{2}{c}{Linear Probing} \\
			      & $k$=1 &  $k$=10 &  $k$=20 & Top-1 & Top-5  \\
			     \midrule
			     10 random negatives & 39.5 & 45.0 & 45.3 & 60.1 & 82.7 \\
			       \rowcolor{almond} top-10 strongest negatives & \textbf{55.3} & \textbf{60.3} & \textbf{60.5} & \textbf{69.2} & \textbf{88.4} \\
			\bottomrule
		\end{tabular}
		}
  \caption{\textbf{Importance of negatives} \label{tab:sampling_neg}}
\end{subtable} \\

\begin{subtable}[t]{0.98\linewidth}
    \centering
        \resizebox{0.8\columnwidth}{!}{
		 \begin{tabular}[t]{cc|ccc}
			    \toprule 
			      \multirow{2}{*}{Method} & \multirow{2}{*}{Space} & \multicolumn{3}{c}{$k$-NN Evaluation} \\
			      & &  $k$=1 &  $k$=10 &  $k$=20 \\
			     \midrule
			    SimCLR & Projection Space & 35.8 & 41.6 & 42.3 \\
			      \rowcolor{almond}  SimCLR & Representation Space & \textbf{46.0} & \textbf{51.5} & \textbf{51.9} \\
                 \midrule
			    GroCo & Projection Space & 51.4 & 56.9 & 57.3\\
			    \rowcolor{almond} GroCo & Representation Space & \textbf{55.3} & \textbf{60.3} & \textbf{60.5}\\
			\bottomrule
		\end{tabular}
        }

  \caption{\textbf{$k$-NN evaluation} \label{tab:knn_eval}}
\end{subtable} \\

\caption{\textbf{Additional Experiments.} The best results are bolded. Options used to obtain the main results are highlighted.
\textbf{Backbone=Resnet50, Views=2$\times$224, \#epochs=100.} \label{tab:ablation-sm}}
  \vspace{-4mm}
\end{table}

{
\myparagraph{Top negatives, stop gradient operation.} 
We further analyze if using the top-$10$ strongest negatives and the stop gradient operation (stop-grad) can also boost the performance of the considered contrastive learning baseline SimCLR.
In our method, the stop gradient operation stabilizes training (fewer spikes in gradients) and allows for convergence with large variations of hyperparameters. 
In Table~\ref{tab:stop_grad}, we observe that stop gradient does not boost SimCLR performance. However, if we utilize only the top-10 strongest negatives in the loss, it stabilizes SimCLR training.
Moreover, usage only top-$10$ negatives indeed boosts SimCLR performance on $+3.1\%$ in $k$-NN ($k=20$) and $+0.3\%$ in linear probing (Top-$1$), but the SimCLR still significantly underperforms the proposed GroCo method by $5.5\%$ in $k$-NN ($k=20$) and $+3.2\%$ in linear probing (Top-$1$).
}

{
\myparagraph{Longer training.} 
We further assess the performance of our model in a longer training regime of $800$ epochs (Table~\ref{tab:longer_training}). 
We find that the proposed model with a multi-crop augmentation strategy achieves $66.1\%$ in $k$-NN ($k=20$) and $73.9\%$ in linear probing (Top-$1$).
}

\myparagraph{Augmentation Strategy.} In Table~\ref{tab:augmentation} we evaluate the performance of the model with respect to different augmentation strategies for view sampling. We follow two setups:  (1) the augmentation strategy as used in the SimCLR~\cite{chen2020simple} method with a random resized crop, color jittering, and gaussian blur, grayscaling and horizontal flip and (2) the augmentation strategy as used in the DINO~\cite{caron2021emerging} method that extends the SimCLR list of augmentations with solarization.
SimCLR augmentations are considered as ``stronger'' compared to DINO augmentations since they include a larger range of cropping sizes ($8\%-100\%$ of original image compared to $14\%-100\%$ in  DINO augmentations) and larger range values in color jittering. We observe that the stronger SimCLR~\cite{chen2020simple} augmentations are more beneficial for the SimCLR method than the weaker DINO augmentations, while for the proposed method, the DINO augmentation strategy is more beneficial. However, the difference between augmentation strategies diminishes with increasing number of training epochs and is no longer measurable at $400$ epochs. For a fair comparison, we use the SimCLR augmentation strategy in all reproductions of the SimCLR method reported in the main paper. 

\myparagraph{Projection Dimensionality.} 
We also ablate our method with respect to dimensionality of the projection space (or the latent space), where distances between samples are computed to calculate a training loss. 
Table~\ref{tab:proj_dim} shows that increasing dimensionality of the projection space increases performance in general, which is more noticeable for the $k$-NN performance. Note that we do not change the dimensionality of the embedding space (output space of the encoder that is used for the $k$-NN evaluation and linear evaluation), which is always 2048-dimensional.

\myparagraph{Importance of Negatives.} 
We also evaluate the importance of utilizing strong negatives for the successful training of our model. We train the model using ten random negatives instead of the top-10 strongest negatives as a negative group and report performance in Table~\ref{tab:sampling_neg}. We observe that leveraging the strongest negatives increases performance across all metrics, demonstrating the importance of hard negatives during training with the GroCo loss, similarly as the contrastive loss benefits from  hard negative sampling~\cite{robinson2020contrastive}.

\myparagraph{$k$-NN in Projection Space.} 
We also evaluate the $k$-NN performance in the projection space (or the latent space) where the training loss is applied. We compare $k$-NN performance in the projection and representation spaces in Table~\ref{tab:knn_eval}. 
We observe that for both methods, $k$-NN performance is higher if we use embeddings from the representation space even though we train the model to compare embeddings in the projection space. 
This could be explained by the fact that the embedding space contains more general image representations since the representations in projection space could be overfitted to the respective augmentations and there become agnostic to some image attributes (like color, since we train the model to match views with different color jittering parameters).

\section{Qualitative Analysis of Learned Representation Space}
\label{sec:repres}

\begin{figure*}[t]
    \centering 
    \begin{subfigure}{0.47\linewidth}
        
        \begin{subfigure}{0.845\linewidth}
            \includegraphics[width=\textwidth]{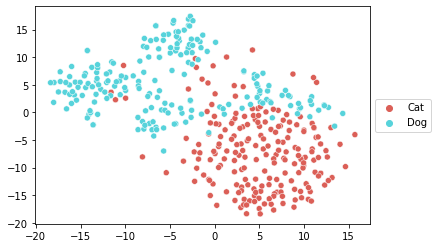}
        \end{subfigure}
        \begin{subfigure}{1\linewidth}
            \includegraphics[width=\textwidth]{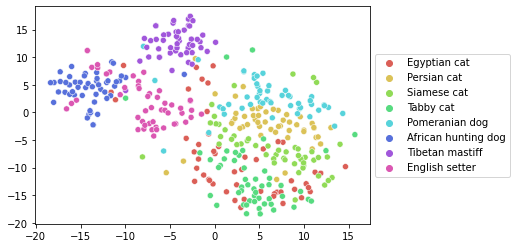}
        \end{subfigure}
        \caption{SimCLR}
    \end{subfigure}
    \hfill
    \begin{subfigure}{0.47\linewidth}
       
        \begin{subfigure}{0.845\linewidth}
            \includegraphics[width=\textwidth]{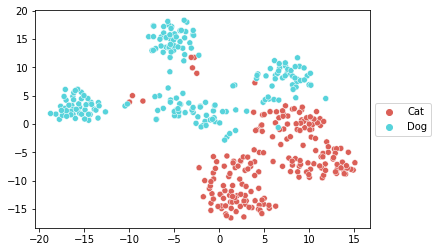}
        \end{subfigure}
        \begin{subfigure}{1\linewidth}
            \includegraphics[width=\textwidth]{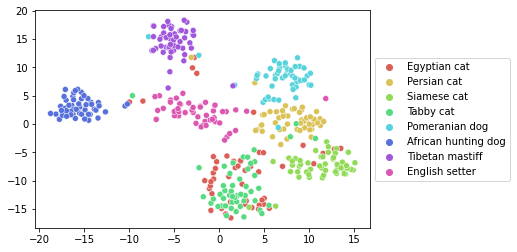}
        \end{subfigure}
        \caption{GroCo (ours)}
    \end{subfigure}
    \caption{t-SNE visualization of learned representations of Imagenet validation images from four classes of different types of cats (Egyptian cat, Persian cat, Siamese cat, Tabby cat) and four classes different types of dogs (Pomeranian dog, African hunting dog, Tibetan mastiff, English setter) for the SimCLR method and the proposed method. For visualization we use models with Resnet50 encoder trained for 100 epochs with a batch size of 1024 and $2 \times 224$ views.
    \label{fig:repres2}}
\end{figure*}

We additionally perform a qualitative analysis of the learned representations. 
In Figure~\ref{fig:repres2}, we visualize representations for images from four classes of different types of cats and four classes of different types of dogs. We find that our method produces much more visually separable clusters with respect to ``inter-class'' variations (cats vs dogs) and ``intra-class'' variations (between different classes of cats) than the SimCLR baseline.

\section{Odd-even Sorting Network}
\label{sec:odd_even}

\begin{figure}[h]
    \centering 
    \begin{subfigure}{\linewidth}
        \centering 
        \includegraphics[width=0.95\textwidth]{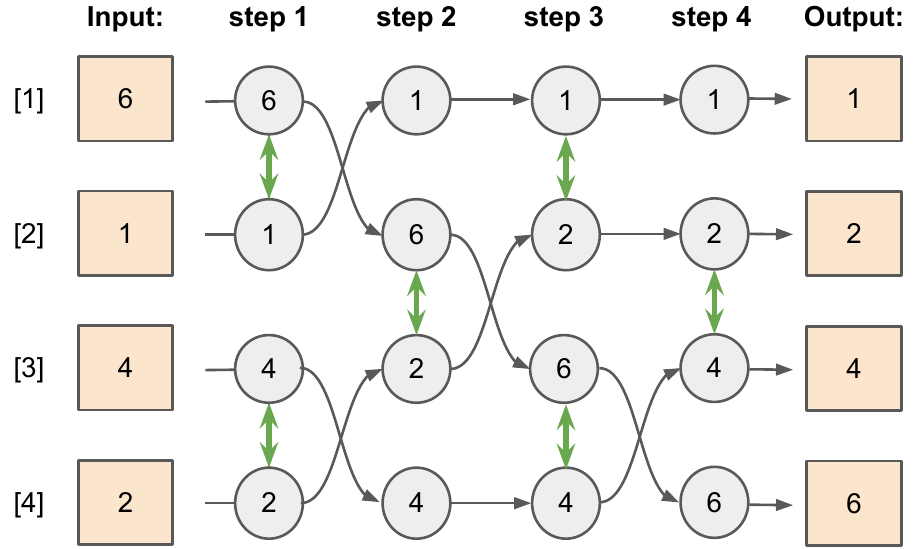}
    \end{subfigure}
    \caption{An illustration of a hard odd-even sorting network for sorting four elements in non-descending order with an example of sorting of [6, 1, 4, 2] array.
    \label{fig:sorting_supplement}}
    \vspace{-2mm}
\end{figure}

An odd-even sorting network, or odd-even sort, is a sorting algorithm from classic computer science literature~\cite{knuth1998sorting}. Sorting networks, or networks for sorting, are a family of sorting algorithms that consist of the \textit{fixed} sequence of comparisons, in a sense that the next comparisons (elements on which positions are compared) does not depend on the result of previous comparisons.  An odd-even sorting network is a simple example of this family of algorithms. The odd-even sorting network compares neighbored
elements starting from odd and even indices alternating on each step, and requires $n$ steps to sort a sequence of $n$ elements. We present a pseudocode of the odd-even sorting network in Algorithm~\ref{algo:odd-even}. Additionally, we illustrate the (hard) odd-even sorting process in Figure~\ref{fig:sorting_supplement}.

\begin{algorithm}[h]
\definecolor{codeblue}{rgb}{0.25,0.5,0.5}
\lstset{
  basicstyle=\fontsize{7.2pt}{7.2pt}\ttfamily\bfseries,
  commentstyle=\fontsize{7.2pt}{7.2pt}\color{codeblue},
  keywordstyle=\fontsize{7.2pt}{7.2pt},
}
\begin{lstlisting}[language=python]
# arr: array to sort
# n: length of array

for s in range(1, n + 1):
    if s %
        for i in range(0, n - 1, 2):
            if arr[i] > arr[i+1]:
                arr[i], arr[i+1] = arr[i+1], arr[i]
    else:
        for i in range(1, n - 1, 2):
            if arr[i] > arr[i+1]:
                arr[i], arr[i+1] = arr[i+1], arr[i]     
\end{lstlisting}
\caption{\small Python--style pseudocode of an odd-even sorting network for sorting an array of numbers in non-descending order}
\label{algo:odd-even}
\end{algorithm}

\vspace{5em}~

\section{Implementation Details}
\label{sec:supp_imp_detail}

\subsection{Linear Evaluation Details}

For linear evaluation, we train a linear classifier on frozen representations in a fully-supervised way, using the training set of ImageNet for training and the validation set for evaluation. 
We follow the training protocol of SimCLR~\cite{chen2020simple} and SimSiam~\cite{chen2021exploring} and train a linear classifier for 90 epochs using the LARS optimizer~\cite{you2017large} with the batch size of $4\,096$, the momentum of $0.9$, the linear rate of $1.6$ (following the rule: $\text{learning rate} = 0.1 \times \text{batch size} / 256$), without a warmup and weight decay. 
Following~\cite{chen2020simple} and \cite{chen2021exploring}, we use weak data augmentation (only random cropping with horizontal flipping) and apply gradient stopping on the input of the classifier to prevent updating the encoder.

\subsection{SimCLR with Multiple Positives}

To train SimCLR with more than one positive view per anchor, we apply contrastive loss for all possible positive pairs, considering all views from other images in the batch as negatives (with a batch of $B$ examples with have  $m (B - 1)$ negatives views). 
Let $x_i^b$ denote the $i$'th view of the $b$'th image in a batch, and $P_{x_i^b}$ denote a set of positive samples for the anchor $x_i^b$, and $N_{x_i^b}$ denote a set of positive samples for the anchor  $x_i^b$.
Then, the loss is calculated as 
\begin{equation}
\begin{aligned}
  &\mathcal{L}_{\text{SimCLR}} =  \frac{1}{B} \sum_{b=1}^B\frac{1}{m}
 \sum_{i=1}^m \frac{1}{\left\lVert P_{x_i^b} \right\rVert}
 \sum_{\substack{y} \in P_{x_i^b}}
 -\log \Biggl( \\
 & {\frac{\exp(- \text{d}(x_i^b,y) / \tau)}
 {\exp(-\text{d}(x_i^b, y)/ \tau) + \sum_{\substack{z \in N_{x_i^b}}}{\exp(-\text{d}(x_i^b, z)/ \tau)} } } \Biggl)
\end{aligned}
\end{equation}
where $\tau$ is a temperature parameter. 
This extension of the SimCLR framework for $m > 2$ views per image is the same as used in the SwAV evaluations~\cite{caron2020unsupervised}. Note that in the multi-crop scenario, we use only full-resolution global
views as positive examples following ``local-global'' correspondence idea~\cite{caron2020unsupervised,caron2021emerging}.

\end{document}